\documentclass[5p]{elsarticle}

\makeatletter
\def\ps@pprintTitle{%
 \let\@oddhead\@empty
 \let\@evenhead\@empty
 \def\@oddfoot{}%
 \let\@evenfoot\@oddfoot}
\makeatother

\usepackage{hyperref}
\usepackage{graphicx}
\usepackage{amsmath}
\usepackage{epstopdf}
\usepackage{amssymb}
\usepackage{textcomp}
\usepackage{siunitx}
\usepackage{url}
\usepackage{subfig}
\usepackage{todonotes}

\journal{Medical Image Analysis}

\biboptions{authoryear}

\begin{document}

\begin{frontmatter}

\title{Coronary Artery Centerline Extraction in Cardiac CT Angiography \\ Using a CNN-Based Orientation Classifier}

\author[isi]{Jelmer M. Wolterink\corref{mycorrespondingauthor}}\ead{j.m.wolterink@umcutrecht.nl}
\author[rad]{Robbert W. van Hamersvelt}\ead{r.w.vanhamersvelt-3@umcutrecht.nl}
\author[isi]{Max A. Viergever}\ead{m.viergever@umcutrecht.nl}
\author[rad]{Tim Leiner}\ead{t.leiner@umcutrecht.nl}
\author[isi]{Ivana I\v{s}gum}\ead{i.isgum@umcutrecht.nl}

\cortext[mycorrespondingauthor]{Corresponding author}

\address[isi]{Image Sciences Institute, University Medical Center Utrecht \& Utrecht University, Q.02.4.45, P.O. Box 85500, 3508 GA Utrecht, The Netherlands}
\address[rad]{Department of Radiology, University Medical Center Utrecht \& Utrecht University, E.01.132, P.O. Box 85500, 3508 GA Utrecht, The Netherlands}

\begin{abstract}
Coronary artery centerline extraction in cardiac CT angiography (CCTA) images is a prerequisite for evaluation of stenoses and atherosclerotic plaque.
In this work, we propose an algorithm that extracts coronary artery centerlines in CCTA using a convolutional neural network (CNN).

In the proposed method, a 3D dilated CNN is trained to predict the most likely direction and radius of an artery at any given point in a CCTA image based on a local image patch. Starting from a single seed point placed manually or automatically anywhere in a coronary artery, a tracker follows the vessel centerline in two directions using the predictions of the CNN. Tracking is terminated when no direction can be identified with high certainty. The CNN is trained using manually annotated centerlines in training images. No image preprocessing is required, so that the process is guided solely by the local image values around the tracker's location.

The CNN was trained using a training set consisting of 8 CCTA images with a total of 32 manually annotated centerlines provided in the MICCAI 2008 Coronary Artery Tracking Challenge (CAT08). Evaluation was performed within the CAT08 challenge using a test set consisting of 24 CCTA test images in which 96 centerlines were extracted. 
The extracted centerlines had an average overlap of 93.7\% with manually annotated reference centerlines. Extracted centerline points were highly accurate, with an average distance of 0.21 mm to reference centerline points. 
Based on these results the method ranks third among 25 publicly evaluated methods in CAT08. In a second test set consisting of 50 CCTA scans acquired at our institution (UMCU), an expert placed 5,448 markers in the coronary arteries, along with radius measurements. Each marker was used as a seed point to extract a single centerline, which was compared to the other markers placed by the expert. This showed strong correspondence between extracted centerlines and manually placed markers. In a third test set containing 36 CCTA scans from the MICCAI 2014 Challenge on Automatic Coronary Calcium Scoring (orCaScore), fully automatic seeding and centerline extraction was evaluated using a segment-wise analysis. This showed that the algorithm is able to fully-automatically extract on average 92\% of clinically relevant coronary artery segments. Finally, the limits of agreement between reference and automatic artery radius measurements were found to be below the size of one voxel in both the CAT08 dataset and the UMCU dataset. Extraction of a centerline based on a single seed point required on average 0.4 $\pm$ 0.1 s and fully automatic coronary tree extraction required around 20 s.

The proposed method is able to accurately and efficiently determine the direction and radius of coronary arteries based on information derived directly from the image data. The method can be trained with limited training data, and once trained allows fast automatic or interactive extraction of coronary artery trees from CCTA images.
\end{abstract}


\end{frontmatter}

\section{Introduction}
Accurate information about the geometry and topology of a patient's vasculature is crucial for many medical applications. In patients with suspected coronary artery disease, information about the cardiac vasculature may be obtained non-invasively  using a cardiac CT angiography (CCTA) scan \citep{Leip14}. A typical first step in the analysis of CCTA scans is the extraction of coronary artery lumen centerlines, which allow multi-planar reconstructions that facilitate stenosis detection and plaque identification. Manual extraction of coronary centerlines is a tedious and time-consuming process, which is infeasible in clinical practice. Therefore, (semi-)automatic methods have been proposed for coronary centerline extraction.

(Semi-)automatic centerline extraction has long been a topic of research in CCTA and other medical images showing vascular structures. A comprehensive review of methods was provided by \cite{Lesa09}. Three main categories of (semi-)automatic centerline extraction methods can be identified. Methods in the first category compute a minimal cost path between start- and end-points that are defined either manually or automatically \citep{Wink02,Kris08}. Minimal cost paths have a high overlap with the reference centerlines, but may suffer from shortcuts between different points on the centerline. Hence, the design of a cost function that is low at centerlines and high at other locations is essential. 
A second category consists of methods that first obtain a segmentation \citep{Stef01,Yang12} or localization \citep{Zhen13} of the coronary artery tree, and subsequently recover the lines at the center of these structures. These methods typically do not require any user input, and are likely to bridge gaps and discontinuities in the arteries, which may appear due to pathology or imaging artifacts. However, they require analysis of the full 3D volume and may be time-consuming. A third category of methods tracks the coronary artery centerline based on iterative determination of the vessel's location, orientation and radius \citep{Aylw02,Frim10,Zhou12,Ceti15,Lesa16}. This leads to very sparse exploration of the CCTA image and low computational overhead. However, due to their sparse nature such methods are more sensitive to gaps, discontinuities and stenoses in the artery.

Regardless of the employed approach, methods for vessel centerline extraction require a filter or model to determine the vessel location, orientation and radius. Commonly used filters are based on eigenvectors of the Hessian matrix \citep{Fran98,Sato98} or idealized tubular models of vessels \citep{Frim10}. More often than not, such filters and models are hand-crafted based on assumptions about the appearance of the coronary arteries and may not fully grasp the information available in the data. Consequently, they require adaptation to cases in which the underlying assumptions do not hold. Such adaptations may be required at coronary branching points, or to suppress responses in non-coronary structures \citep{Yang12}. Explicit modeling of all exceptions is a challenging task.

As an alternative to hand-crafted vesselness filters, machine learning has recently been used to learn vessel models from annotated data. \cite{Schn15} proposed to identify the location of the centerline using steerable features and randomized decision trees, while \cite{Siro15} trained a boosting regressor to predict, for each voxel, the proximity to the closest centerline. \cite{Guls16} used a support vector machine (SVM) classifier to learn an orientation flow field in CCTA based on steerable features and spatial features extracted using automatically identified anatomical landmarks. In all these methods, the final centerline was extracted by finding a minimum energy or maximum flow path in the filtered image. Although these methods do not depend on hand-crafted vessel filters or models, they require an intermediate feature representation of the data and evaluation at multiple voxel locations to obtain the vessel orientation at a single point.

Convolutional neural networks (CNNs) have recently demonstrated the ability to derive useful features from image data in a wide range of medical image analysis tasks \citep{Litj17}. Specifically, they have shown the ability to learn data representations for vessel segmentation in retinal fundus images \citep{Wang15} or cardiac CT \citep{Moes16b,Wu16}. 
This suggests that CNNs could also be used to determine vessel centerlines. 
In this work, we propose a CNN that learns to identify the coronary centerline direction and lumen radius directly from image data only. Hence, all information is extracted directly from the image, and no intermediate hand-crafted vesselness representations are required. 
The method can be trained with manually annotated reference centerlines in a limited set of training images. 
We describe an algorithm that utilizes the direction and radius predictions by the CNN to extract centerlines based on a single seed point that can be placed either manually or automatically. We evaluate the performance of the proposed method using the publicly available Rotterdam Coronary Artery Evaluation Framework\footnote{\url{http://coronary.bigr.nl/centerlines}}, a set of CCTA scans acquired at our institution, and a set of CCTA scans acquired using two additional CT scanners. Our experiments demonstrate that the trained CNN allows fast and accurate centerline extraction. This allows automatic or interactive extraction of the coronary artery tree in a matter of seconds. 

\section{Data}
\label{sec:data}
The method was trained and evaluated using CCTA scans from a coronary centerline extraction challenge (CAT08), scans obtained at our institution (UMCU), and scans obtained using two additional CT scanners (orCaScore).

\subsection{CAT08 dataset}
We included all 32 CCTA images from a publicly available evaluation framework for coronary centerline extraction \citep{Scha09}, the MICCAI 2008 Coronary Artery Tracking Challenge (CAT08) which is part of the Rotterdam Coronary Artery Evaluation Framework. Images in CAT08 were acquired on a 64-slice CT scanner (Sensation 64, Siemens Medical Solutions, Forchheim, Germany) or dual-source CT scanner (Somatom Definition, Siemens Medical Solutions, Forchheim, Germany). Scans were acquired with 120 kVp and a maximum tube current of 900 mA. Images were reconstructed to a mean voxel size of 0.32 $\times$ 0.32 $\times$ 0.4 mm$^3$. In each scan, the centerline and radius of four major coronary arteries were manually annotated in a consensus reading by three experts. These arteries were the left anterior descending (LAD), left circumflex (LCX) and right coronary artery (RCA), and a fourth vessel selected as a side-branch of the LAD, LCX or RCA. The CAT08 organizers separated the 32 CCTA images into a training set and a test set based on subjective image quality (6 poor, 11 moderate, 15 good) and coronary calcium burden (12 low, 16 moderate, 4 severe). The training set consists of 8 CCTA images with reference annotations for the centerline location and radius, and the test set consists of 24 CCTA images for which no reference annotations are provided. A detailed description of scan acquisition and reconstruction, and the centerline annotation protocol is provided in \citep{Scha09}.

\subsection{UMCU dataset}
We included 50 CCTA scans that were consecutively acquired at our institution (UMC Utrecht, Utrecht, The Netherlands). The need for informed consent was waived by the local Medical Ethics Committee, due to the retrospective nature of this study. These scans were acquired with a 128-detector row Philips Brilliance iCT (Philips Healthcare, Best, The Netherlands) scanner, with contrast enhancement and ECG-triggering, using 120 kVp and 210-300 mAs. Images had a slice spacing of 0.45 mm and an average in-plane resolution of 0.44 $\pm$ 0.04 $\times$ 0.44 $\pm$ 0.04 mm$^2$.
The quality of these images was generally good: intravascular stents were present in 2/50 scans, step reconstruction artifacts in 7/50 scans, and coronary motion artifacts in 11/50 scans. Furthermore, 26/50 scans contained coronary artery calcification. For patients with coronary artery calcification (CAC), Agatston calcium scores in corresponding non-contrast CT images ranged from 0.5 to 3,540, with a median (IQR) of 31.6 (6.3--610.5) \citep{agat90a}.

In each of the 50 scans, an expert observer manually placed markers in the coronary arteries separated by approximately 10 mm, along with a measurement of the vessel radius in the axial plane at the marker location. Markers were placed in all visible major coronary arteries and branches. The radii of the placed markers ranged from 0.45 mm to 3.51 mm, with a median (IQR) of 0.81 (0.61--1.16) mm. In total, 5,448 markers were placed along the coronary artery centerlines with an average of 109 markers per CCTA scan. In addition, the location of the left and right coronary ostium were manually indicated.

\begin{figure}
\centering
\includegraphics[width=\linewidth]{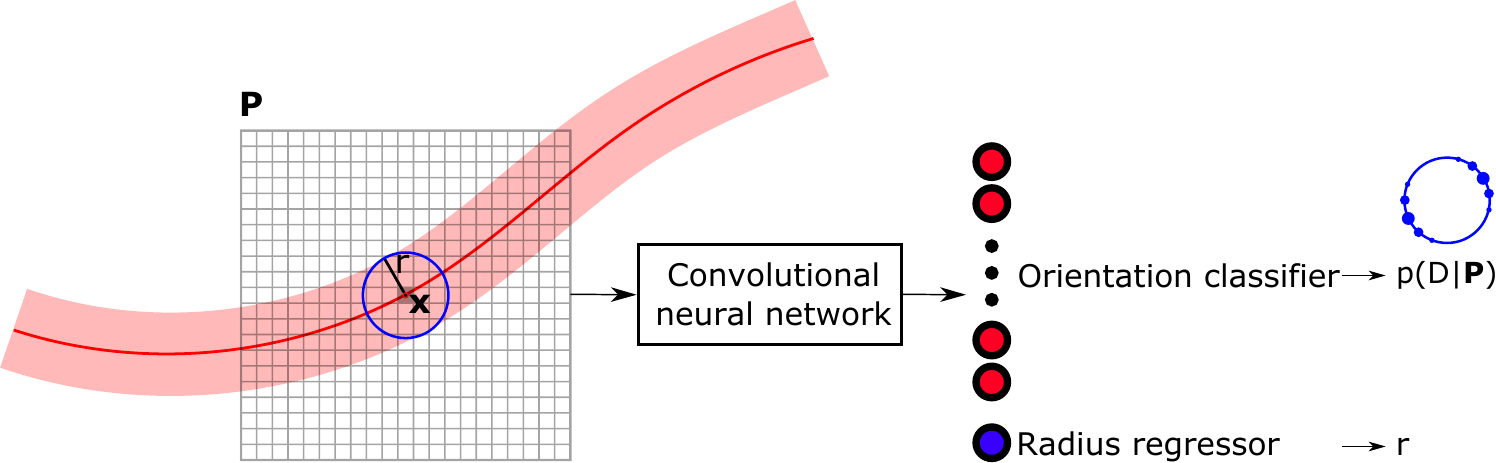}
\caption{Overview of the proposed method. At location $\mathbf{x}$, an isotropic 3D patch $\mathbf{P}$ is extracted and used as input to a convolutional neural network (CNN). This CNN simultaneously determines a probability distribution $p(D|\mathbf{P})$ over a discrete set of directions on a sphere (here shown as a blue circle), and an estimate $r$ of the radius of the vessel.}
\label{fig:methodoverview}
\end{figure}

\subsection{orCaScore dataset}
We included 36 CCTA scans that were provided in the MICCAI 2014 Challenge on Automatic Coronary Calcium Scoring (orCaScore) \citep{Wolt16b}. This set includes 18 images that were acquired on a GE Lightspeed VCT scanner (GE Healthcare, Milwaukee, Wisconsin) and 18 images that were acquired on a Toshiba Aquilion ONE scanner (Toshiba Medical Systems, Otawara, Japan). All scans were acquired with contrast-enhancement and ECG-triggering. On each scanner, nine male and nine female patients were scanned. Patient inclusion was stratified based on a commonly used CAC score categorization (I:0, II:1-100, III:101-300, IV:\textgreater 300). We included eight patients in categories I and III, and ten in categories II and IV. For a detailed description of this dataset, please see \citep{Wolt16b}. For this work, the locations of the left and right coronary ostium were manually indicated in each image.

\section{Method}
\label{sec:method}
We propose to determine the orientation and radius of a coronary artery at a location $\mathbf{x}$ in an image $I$, based on a 3D isotropic image patch $\mathbf{P}$ and using a single CNN (Fig. \ref{fig:methodoverview}). The output layer of the CNN consists of classification nodes that determine a posterior probability distribution over possible tracking directions $D$ and a regression node that determines the radius $r$ of the vessel at point $\mathbf{x}$. 

\subsection{Convolutional neural network}
\label{sec:architecture}
The proposed CNN takes a 3D image patch $\mathbf{P}$ of $w \times w \times w$ voxels with  isotropic voxel spacing $v$ (in mm) centered at $\mathbf{x}$ as input, and determines a posterior probability distribution $p(D|\mathbf{P})$ over a discrete set of possible directions $D$, as well as an estimate  $r$ of the radius. The values for $w$ and $v$ together determine the input resolution and physical receptive field of the CNN in world coordinates, and may be chosen depending on vessel caliber. Here, $w=19$ and $v=0.5$, corresponding to a physical receptive field of 9.5 mm that is sufficient to cover even the widest coronary arteries and their context \citep{Dodg92,Medr16}. 

\begin{figure}[tp!]
\centering
\includegraphics[width=\linewidth]{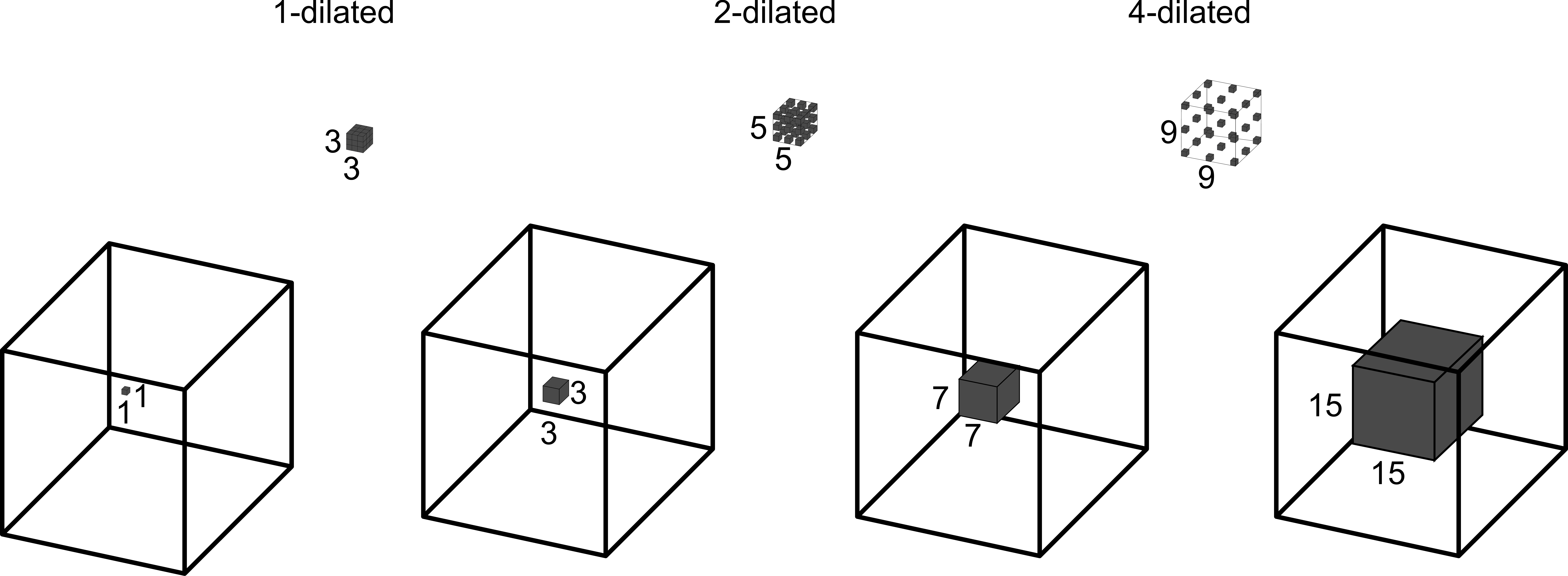} 
\caption{Stacked application of 3D convolution kernels with increasing levels of dilation. A $3\times 3\times 3$ voxel kernel is shown with a stride of 1, 2 or 4 voxels between kernel elements. These kernels rapidly increase the receptive field from $1\times 1\times 1$ voxel to $15\times 15\times 15$ voxels using only 27 trainable parameters per kernel.}
\label{fig:dilate3D}
\end{figure}

The CNN architecture contains a stack of dilated convolution layers that aggregate features over multiple scales with convolution kernels that have an increasing level of dilation, i.e. increased strides between the kernel elements  \citep{Yu15}. Fig. \ref{fig:dilate3D} shows the effect of stacking $3\times 3 \times 3$ voxel 3D convolution kernels with increasing levels of dilation (1, 2 and 4). As the level of dilation increases, the receptive field at each layer increases as well, but the number of trainable parameters per kernel stays the same at $3 \times 3 \times 3 = 27$. Hence, the receptive field grows exponentially from 3 to 7 to 15, but the number of trainable parameters increases linearly. Reducing the number of parameters could prevent overfitting of 3D CNNs and allows fast processing. In addition, the proposed CNN architecture uses full resolution feature maps throughout its architecture instead of downsampling layers. This  makes the network applicable to any image of size $\geq w$.

\begin{table}[tp!]
\centering
\caption{Architecture of the convolutional neural network (CNN) for a $19\times 19\times 19$ voxel 3D input patch. For each layer (Layer), the convolution kernel width (Kernel width) is listed, as well as the dilation level (Dilation), the number of output channels (Channels), and the receptive field at that layer (Field width). All operations are performed in 3D. The number of output channels is equal to the number of potential directions in $D$, plus one channel for radius estimation.}
\label{tab:network}
\begin{tabular}{c|c|c|c|c|c|c|c}
Layer        & 1  & 2  & 3  & 4  & 5  & 6  & 7     \\ \hline
Kernel width & 3  & 3  & 3  & 3  & 3  & 1   & 1   \\
Dilation     & 1  & 1  & 2  & 4  & 1  & 1   & 1   \\
Channels     & 32 & 32 & 32 & 32 & 64 & 64 & $|D|$+1 \\
Field width  & 3  & 5  & 9  & 17 & 19 & 19  & 19 
\end{tabular}
\end{table}

\begin{figure}[t!]
\hfill
\centering
\subfloat[]{
\includegraphics[width=0.45\linewidth]{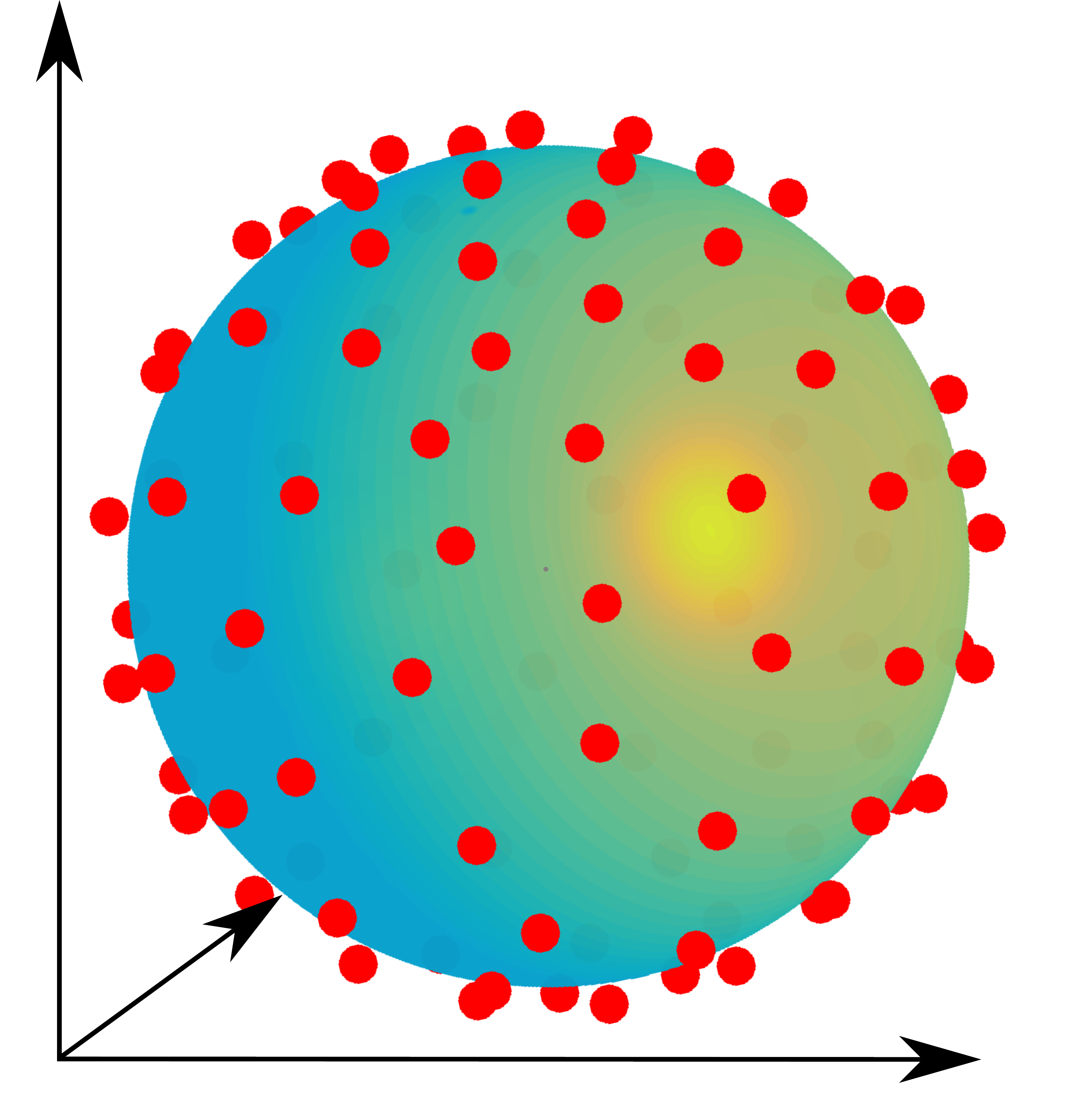} 
\label{subfig:orientations}
}
\hfill
\subfloat[]{
\includegraphics[width=0.45\linewidth]{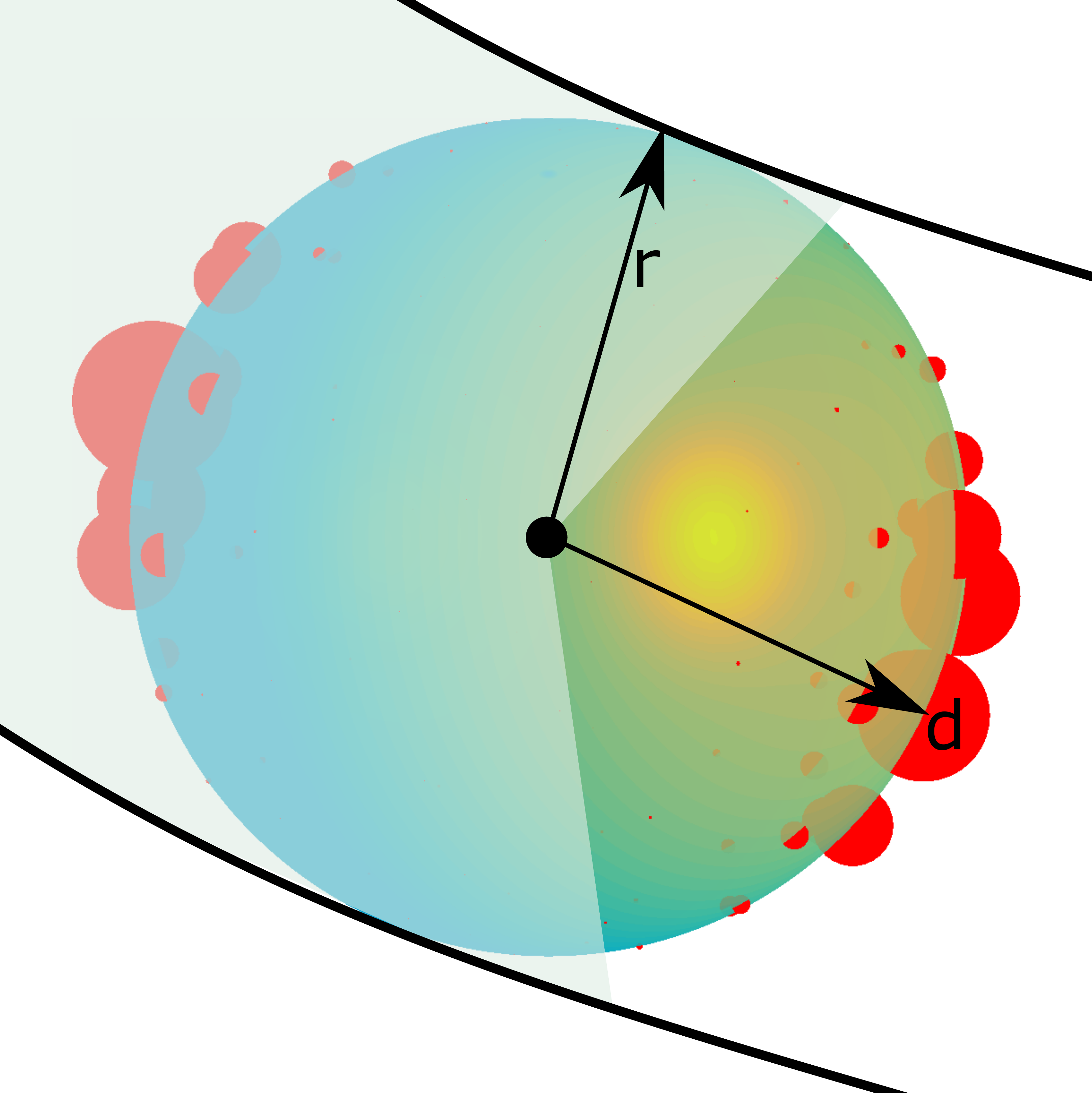}
\label{subfig:activations}
}
\hfill
\caption{\protect\subref{subfig:orientations} The set of possible directions $D$ is distributed on a sphere. \protect\subref{subfig:activations} During testing, a posterior probability distribution over $D$ is determined, and the tracker follows the direction $d$ corresponding to the maximum in this distribution. Only directions with an angle $\leq60\si{\degree}$ to the previously followed direction are considered.}
\label{fig:sphere}
\end{figure}

Table \ref{tab:network} lists characteristics of the proposed CNN architecture; the width of convolution kernels, the level of dilation, the number of output channels, and the receptive field at each layer. 
No dilation is applied in the first two or final three layers of the network. To convey more information about an increasing receptive field, the network gets wider towards the output layers. 
The output layer combines two tasks: direction classification and radius regression. The possible directions $D$ are distributed on a sphere, where each point corresponds to a class (Fig. \ref{subfig:orientations}). Orientation determination is posed as a classification problem rather than a regression problem, so that the CNN may return a posterior probability distribution with multiple local maxima during tracking (Fig. \ref{subfig:activations}). In contrast, a regression model minimizing the squared error between the predicted and reference direction would predict the average direction, which in many cases is the center of the sphere. 
The $|D|$ classification nodes are combined through a softmax activation layer. In contrast to the potential directions, the radius can be estimated by a single scalar value. Hence, this value is estimated using regression with a linear activation function. The $|D|$ classification nodes and the single regression node together form the output layer of the CNN. 

Aside from the softmax classification and linear regression nodes in the output layer, all nodes in the network use rectified linear units (ReLUs) as activation function. Batch normalization is applied in each layer of the network \citep{Ioff15}. Fully connected layers are implemented as $1 \times 1 \times 1$ convolutions. Hence, after training, the network may efficiently be applied to images of arbitrary size.

\begin{figure*}[htp]
\centering
\subfloat[Original training sample]{
\includegraphics[width=0.30\linewidth]{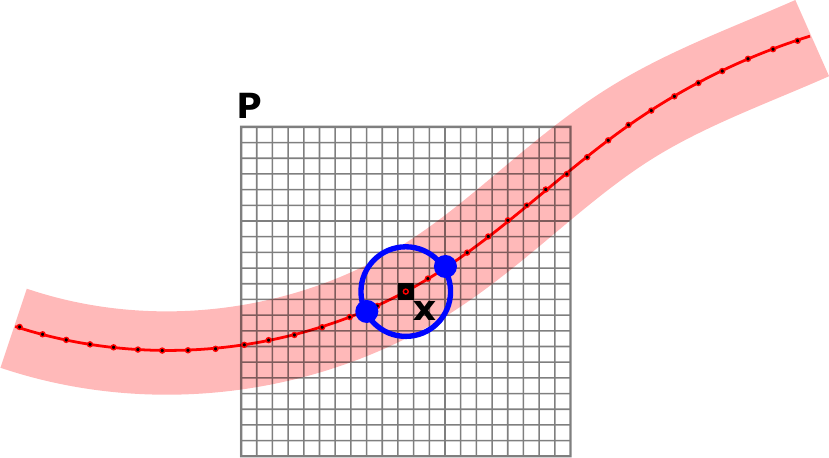}
\label{subfig:standard}
} 
\hfill
\subfloat[Translated training sample]{
\includegraphics[width=0.30\linewidth]{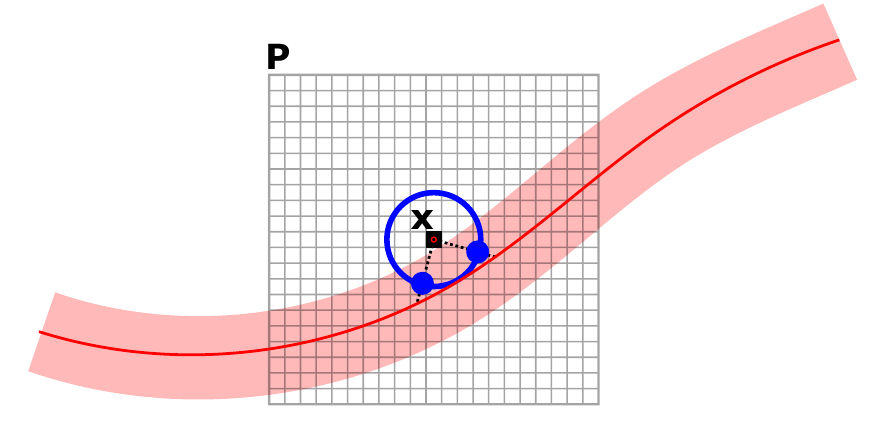}
\label{subfig:translate}
} 
\hfill
\subfloat[Rotated training sample]{
\includegraphics[width=0.30\linewidth]{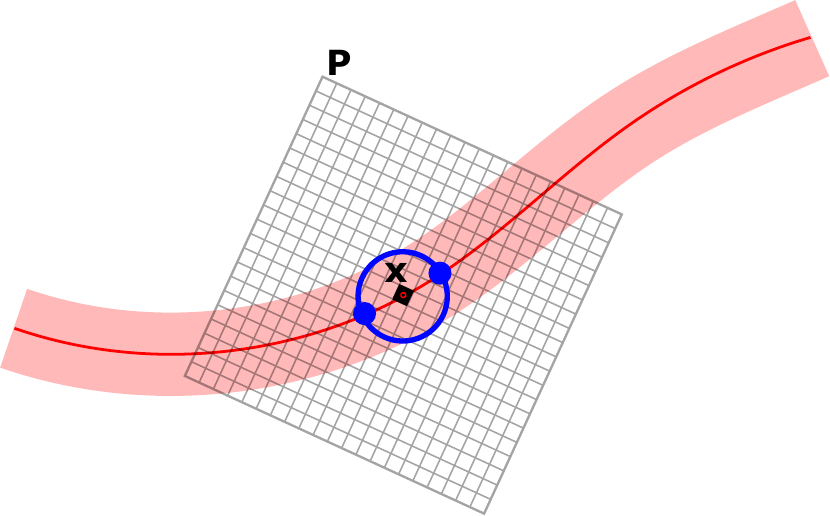}
\label{subfig:rotate}
}
\caption{Training sample extraction for the convolutional neural network (CNN). \protect\subref{subfig:standard} Standard training sample extracted at reference point $x$, consisting of a patch $\mathbf{P}$ and two reference directions in $D$ (shown in blue). Manually defined reference points are shown on the centerline. During training, a random combination of translation and rotation augmentations is applied to each sample. \protect\subref{subfig:translate} A translated off-centerline point $x$ with corresponding patch $\mathbf{P}$ and reference directions. \protect\subref{subfig:rotate} A centerline point $x$ with a rotated patch $\mathbf{P}$ and reference directions.}
\label{fig:training}
\end{figure*}

\subsection{Training strategy}
\label{sec:training}
The CNN is trained with 3D CCTA images from the CAT08 challenge. In each of these images, reference annotations are provided for four centerlines. These reference annotations consist of an ordered set of centerline points with corresponding radius measurements. Not all coronary artery branches are manually annotated, and exhaustive manual annotation of all vessel centerlines in a training image would be a time-consuming and tedious process.  
To only train the CNN with samples for which we can set a reference label with sufficient certainty, we sample training points in those areas where the coronary centerlines have been annotated. 

To sample a training point, a location $\mathbf{x}$ along the reference centerline is randomly selected and a 3D image patch $\mathbf{P}$ centered at that point is extracted from the training image (Fig. \ref{subfig:standard}). The reference radius $r$ at $\mathbf{x}$ is taken from the reference standard, and reference directions are determined as follows. First, a point $\mathbf{x'}$ is found at a distance $r$ from $\mathbf{x}$ along the reference centerline. The displacement vector $\mathbf{\Delta_x}$ between $\mathbf{x}$ and $\mathbf{x'}$ is used to determine a reference direction. For this, direction class $d\in D$ (Fig. \ref{subfig:orientations}) with the smallest angle to $\mathbf{\Delta x}$ is assigned as the reference direction. This direction is considered to be one of two reference directions at point $\mathbf{x}$ and its class probability is set to 0.5 in the reference distribution over $D$. This process is repeated for the opposite direction of the centerline. Hence, two direction classes in the reference distribution over $D$ have probability 0.5 and all other direction classes have probability 0.0. The patch $\mathbf{P}$, the reference distribution over $D$, and the reference radius $r$ together form an input sample for the CNN.

Training with a large and diverse set of training samples is likely to improve the performance of the proposed supervised machine learning method. However, obtaining reliable reference centerlines in a large number of  images is tedious and time-consuming, and may require consensus of multiple experts \citep{Scha09}. Hence, we augment available reference centerline annotations in two ways.

First, we include samples that are located off the coronary artery centerline. 
If the CNN would be only trained with samples that are exactly on the coronary artery centerline, the CNN may provide incorrect predictions when the tracker finds itself slightly off the centerline, causing the tracker to deviate. Consequently, the tracker may be unable to recover. 
Like on-centerline samples, off-centerline samples are extracted using the reference annotations. A point $\mathbf{x}$ on a reference centerline is randomly sampled. To then obtain an off-centerline sample, point  $\mathbf{x}$ is translated using a random shift sampled from a 3D normal distribution with $\mu=0.0, \sigma=0.25r$ (Fig. \ref{subfig:translate}). To determine the reference direction from $\mathbf{x}$, we first identify the closest point on the reference centerline and from this point find $\mathbf{x'}$ at a distance $r$ along the centerline. Then the displacement vector $\Delta\mathbf{x}$ and reference direction is determined as before, based on $\mathbf{x}$ and $\mathbf{x'}$. This process is repeated for the opposite centerline direction.

Second, we enrich the dataset by applying random rotations to input patches (Fig. \ref{subfig:rotate}).
Each training patch $\mathbf{P}$ is rotated around the $x$-, $y$-, or $z$-axis with a random angle $\theta\in [0,2\pi]$.
This balances the orientation of vessels in the training set and makes the CNN agnostic to the orientation of the image.

During training, the Adam optimizer \citep{King14} updates the network parameters $\theta$ to minimize the loss 
\begin{equation}
\ell(\theta) = \ell_c(\theta) + \lambda_r\ell_r(\theta) + \lambda_w||\theta||^2, 
\end{equation}

where $\ell_c$ is the categorical cross-entropy between the reference and posterior probability distributions over the direction class set $D$, $\ell_r$ is the squared error regression loss between the reference and predicted radius values, weighted by a parameter $\lambda_r$, and $\lambda_w||\theta||^2$ is a regularization term on the network parameters. To balance the contribution of each term to the loss function, we used $\lambda_r=10$ and $\lambda_w=0.001$ throughout our experiments. Mini-batch training is used with batches containing 64 randomly selected samples. We used learning rate decay, where the learning rate started at 0.01 and was reduced by a factor 10 every 10,000 iterations, for a total of 50,000 iterations.

\subsection{Iterative tracking}
\label{sec:iterative}
Due to its purely convolutional architecture, the CNN is able to process input images of any size. Hence, the CNN may be integrated in a number of existing methods for coronary centerline extraction, including those that preprocess full CCTA images or those that only sparsely explore the image. We here demonstrate the ability of the CNN to extract centerlines using an iterative tracking algorithm.

The tracking algorithm starts at a seed point $\mathbf{x}_0$. An isotropic 3D patch $\mathbf{P}_0$ centered at $\mathbf{x}_0$ is extracted and processed by the CNN. The output of the CNN consists of the posterior probability distribution $p(D|\mathbf{P}_0)$ over possible directions, and the estimated radius value $r_0$. To determine two initial opposing directions of the tracker, two local maxima $d_0$ and $d'_0$  separated by an angle $\geq90\si{\degree}$ are identified in $p(D|\mathbf{P}_0)$. The tracker will first follow the centerline in the direction $d_0$ until termination, and then in the direction $d'_0$ until termination.

To follow the centerline in the direction $d_0$, the tracker takes a step of length $r_0$ towards $d_0$ and arrives at point $\mathbf{x}_1$. Hence, the step size depends on the local radius of the vessel. Then, a new patch $\mathbf{P}_1$ is extracted at $\mathbf{x}_1$ and processed by the CNN to provide $p(D|\mathbf{P}_1)$ and $r_1$. The new tracking direction $d_1$ is selected as the direction with the highest probability in $p(D|\mathbf{P}_1)$. To prevent the tracker from moving backwards, directions that have an angle $\geq60\si{\degree}$ to $d_0$ are excluded from this selection (Fig. \ref{subfig:activations}).
This process is repeated until a stopping criterion is fulfilled. Subsequently, the tracker follows the same process in the direction $d'_0$, starting again at point $\mathbf{x}_0$.

Termination of the tracker is guided by a stopping criterion based on the uncertainty of the direction classifier. At each point along the extracted centerline, the normalized entropy $H(p(D|\mathbf{P})) \in [0, 1]$ of the posterior probability distribution is computed as

\begin{equation}
	H(p(D|\mathbf{P})) = \frac{\sum_{d \in D} - p(d|\mathbf{P}) \log_2 p(d|\mathbf{P})}{\log_2|D|}.
\end{equation}

\noindent The tracker terminates if the entropy of the selected probability distribution crosses a threshold value $\theta_H=0.9$. 
This may happen when the tracker encounters the coronary ostium or the end of the coronary artery. However, stenotic areas, areas with low image contrast or locations affected by stepping artifacts may also lead to high entropy values. To encourage tracking through such areas, the termination entropy is determined as a moving average over the past three steps, similarly to the probabilistic tracking scheme proposed by \cite{Wang12}. In addition, to prevent the tracker from following a path that overlaps with its already extracted centerline, tracking is terminated when the tracker is at a point which is too close to the already tracked centerline. Finally, a simple postprocessing step is applied to correct for potential leakage into adjacent non-coronary vessels. Long centerlines are shortened to a length of 275 mm and pruned to the point where they have the smallest estimated radius value. 

\subsection{Fully automatic coronary tree extraction}
\label{sec:automatic}
The single-seed iterative tracker that we propose requires a seed point in the coronary arteries. While such seed points can be placed manually by a trained expert, this may be prohibitive when extracting trees in a large number of CCTA scans. We therefore propose an extension of the algorithm presented in Sec. \ref{sec:iterative}. This extension uses a CNN to automatically identify seed points for tracker initialization, and a second CNN to automatically locate points at the coronary ostia for identification of successfully tracked centerlines.

The CNN architecture used is almost identical to the architecture used for orientation and radius estimation described in Sec. \ref{sec:architecture}, Table \ref{tab:network}. The only difference is in the output layer: instead of combining classification and regression, the final layer only performs regression. The value that the CNN predicts is the proximity of the center voxel of the input to a coronary artery centerline or coronary ostium. During training, the squared error regression loss between the reference and predicted proximity values is minimized. Reference values are determined using the formulation proposed by \cite{Siro15}. At a location $\mathbf{x}$, the reference proximity value $d(\mathbf{x})$ is defined as 

\begin{equation}
d(\mathbf{x}) = \left\{
\begin{array}{ll}
e^{a\left(1-\frac{\mathcal{D}_C(\mathbf{x})}{d_M}\right)}-1 & \text{if } \mathcal{D}_C(\mathbf{x}) < d_M \\
0 & \text{otherwise,} 
\end{array}\right.
\end{equation}

where $a$ is a constant that we set to $a=6$ as in \citep{Siro15}, $\mathcal{D}_C(\mathbf{x})$ is the distance of point $\mathbf{x}$ to the nearest centerline, and $d_M$ is a distance cutoff value. We train one network to identify the coronary arteries and one network to identify the ostia and set $d_M$ to 4 mm for centerline points and 16 mm for the ostia. For both seed point and ostia detection, the network is trained fully convolutionally with 3D volumes resampled to isotropic 1.0 mm resolution. Training is performed using the same settings as for the centerline extraction CNN.

After processing by the CNN, a predefined number of seed points for tracker initialization are identified as local maxima in the predicted proximity map. Similarly, two local maxima in the proximity map for the ostia are identified as the coronary ostia.  Seed points are placed in a queue and iteratively used to extract single centerlines using the method described in Sec. \ref{sec:iterative}. If a centerline reaches one of the ostia points, it is added to the tree and all queued seed points that the extracted centerline overlaps with are removed from the queue. If a centerline does not reach one of the coronary ostia points, it is discarded. After processing, a number of centerlines remain, which are all connected to the ostia and branch out towards the distal points of the coronary artery tree.

\section{Experiments and results}
\label{sec:results}

The algorithm was implemented in Python using PyTorch \citep{Pasz17}. Experiments were performed using an NVIDIA Titan Xp GPU. Training of a network with 50,000 iterations required 1.5 hours. The processing time for extraction of a single vessel was on average 0.4 $\pm$ 0.1 s. Automatic identification of seeds and ostia locations for tracking both required around 1 s per patient and fully automatic extraction of the coronary artery tree took around 20 s per patient.

\begin{figure*}
\centering
\includegraphics[width=1.0\textwidth]{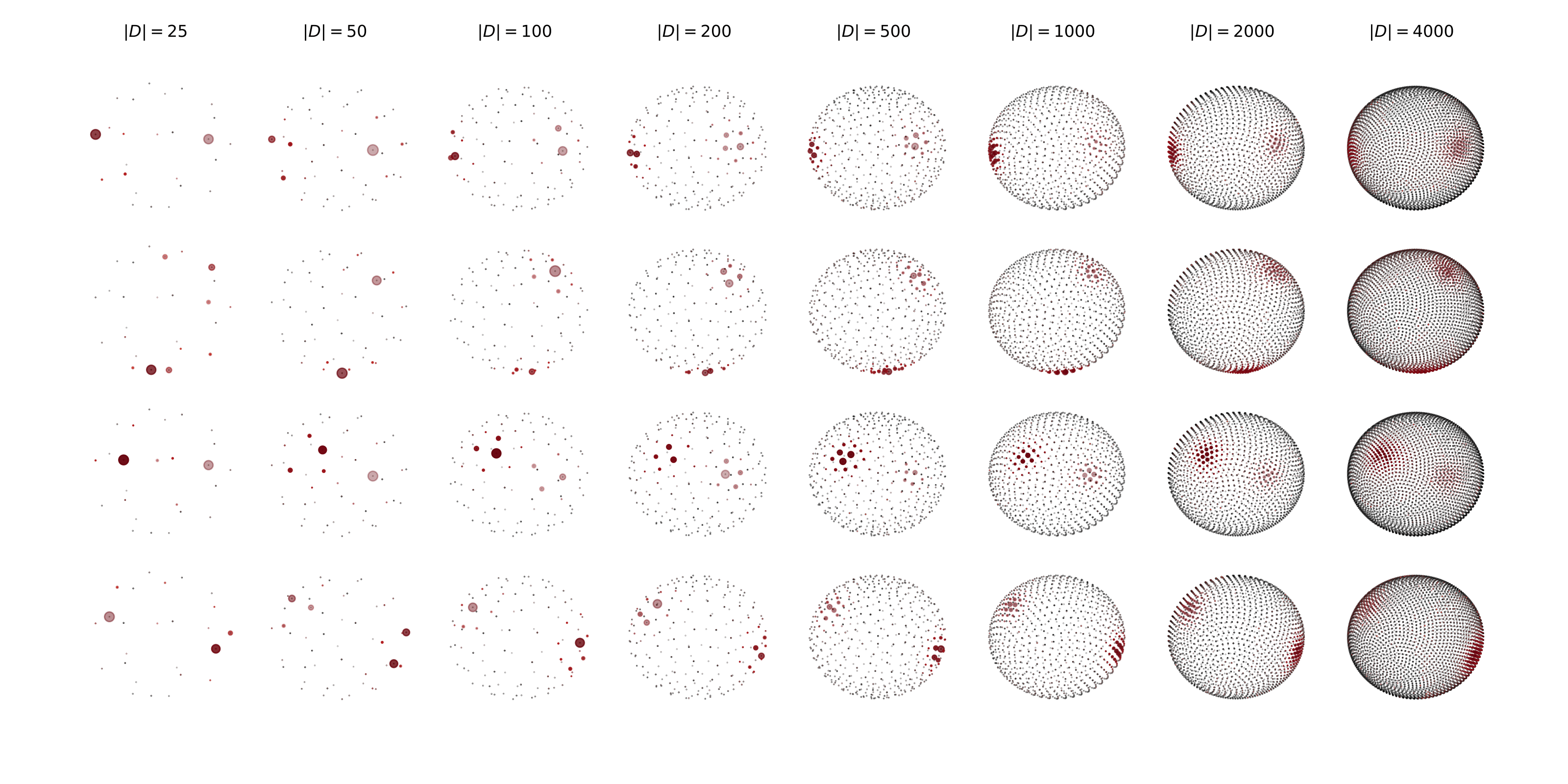}
\caption{Response over the set of orientations $D$, for different numbers of orientations $|D|$. Each row contains responses taken at exactly the same point along the coronary artery, for a total of four points in one vessel.}
\label{subfig:orientationsresponse}
\end{figure*}

\subsection{Centerline extraction}
The method's ability to extract coronary artery centerlines was evaluated in several ways. First, a quantitative analysis of single vessel centerline extraction was performed using the CAT08 challenge data (Sec. \ref{sec:cat08centerline}). The reference centerline annotations in this challenge allow an evaluation of centerline overlap and accuracy. However, the CAT08 challenge only evaluates centerlines for four coronary arteries per patient. Therefore, a second analysis was performed in which single centerlines were extracted based on 5,448 manually placed seed points in 50 CCTA scans acquired at our institution (Sec. \ref{sec:resultsumcu}). This allows evaluation of the robustness of the method in a larger dataset. Third, fully automatic seeding and centerline extraction was evaluated based on the completeness of coronary artery trees (Sec. \ref{sec:orcaresults}). For this, 36 images from the orCaScore challenge were used. While the evaluations in Sec. \ref{sec:cat08centerline} and \ref{sec:resultsumcu} were performed on CCTA scans acquired with Siemens and Philips scanners, images in Sec. \ref{sec:orcaresults} originated from GE and Toshiba scanners. This allows an analysis of the generalization of the method to CCTA scans acquired on different scanners.

\subsubsection{Centerline extraction accuracy}
\label{sec:cat08centerline}
We evaluated centerline extraction performance on the eight training scans provided in CAT08 using leave-one-out cross-validation experiments. For each of the eight images, a CNN was trained using the other seven images. For each vessel, centerline extraction was initialized at a point that uniquely identifies that coronary artery. These points were provided in the CAT08 challenge.

In the CAT08 challenge, extracted centerlines are evaluated based on overlap and accuracy \citep{Scha09}. Total overlap (OV), overlap until first error (OF), and overlap of the extracted centerline with the clinically relevant part of the vessel (radius $\geq 0.75$ mm, OT) are computed using true positive (TP), false positive (FP) and false negative (FN) detections. A TP point lies within the radius of the closest reference point, a FP point does not lie within the radius of any reference point and a FN point is a reference centerline point for which there is no corresponding automatically extracted point. The average inside accuracy metric (AI) measures the average distance between the reference and extracted centerline for automatically extracted points that are within the radius of the reference centerline. Hence, this metric is to some extent independent of the overlap of extracted centerlines.

\begin{figure}
\includegraphics[width=\linewidth]{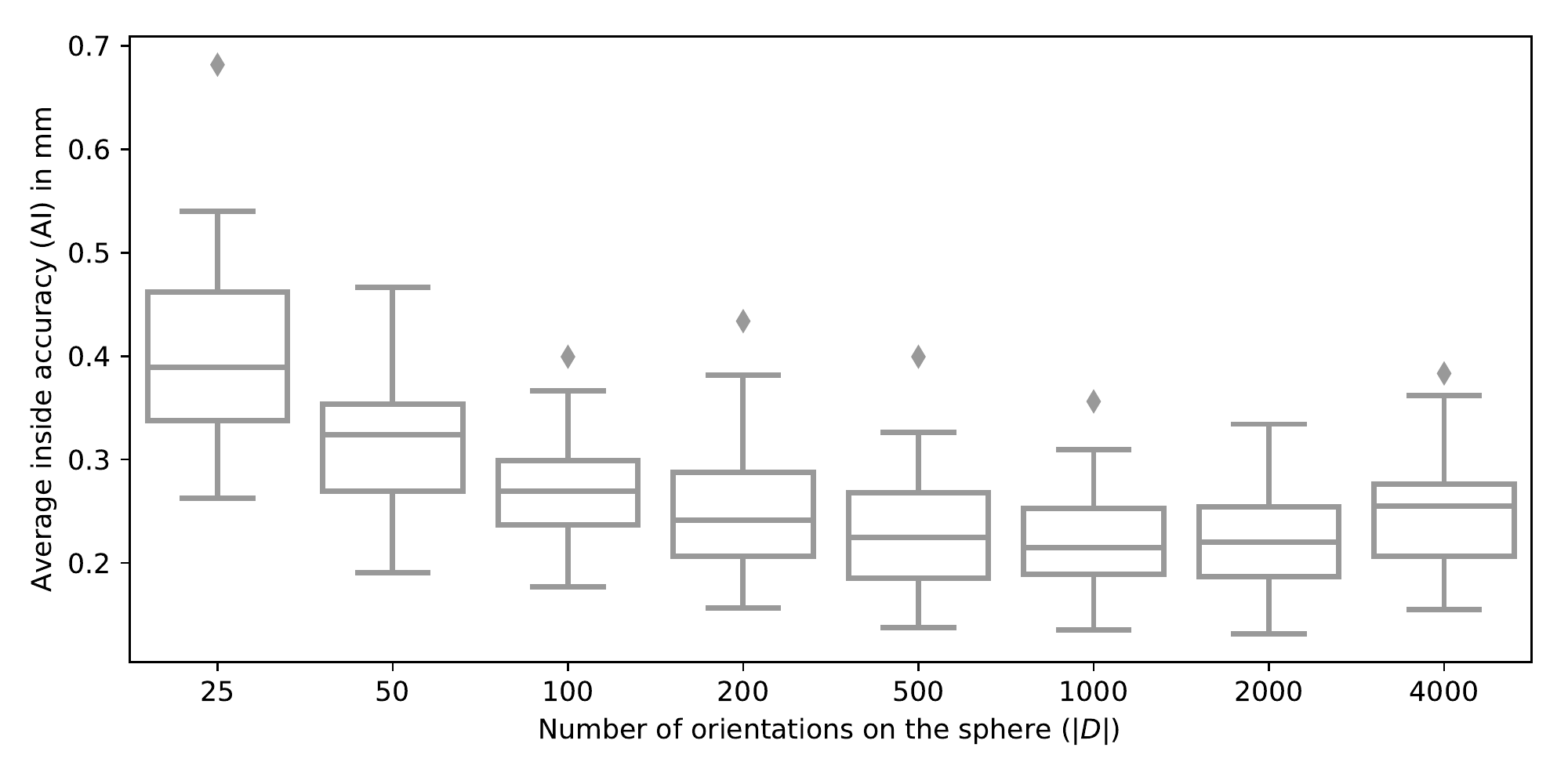}
\caption{The average inside accuracy (AI, lower values are better) of 32 automatically extracted centerlines in the CAT08 training set, obtained using different numbers of orientations $|D|$ on the sphere.}
\label{subfig:aiboxplots}
\end{figure}

To identify the optimal number of possible vessel orientations on the sphere, we first performed experiments set using spheres with $|D|=\{25, 50, 100, 200, 500, 1000, 2000,$ $ 4000\}$ orientations. Fig. \ref{subfig:orientationsresponse} shows example responses of the CNN for different numbers of directions $|D|$, taken at four points along a reference centerline in the CAT08 training set. This shows that larger values of $|D|$ lead to more fine-grained responses, but also to more noisy responses. A quantitative analysis using the AI accuracy between automatically extracted centerlines and reference centerlines in the CAT08 training set shows that the best results are obtained for $|D|=\{500,1000,2000\}$ (Fig. \ref{subfig:aiboxplots}). In the remainder of this work, we use $|D|=500$.

\begin{table}
\centering
\caption{Results for centerline extraction in the CAT08 \textit{training} set. For each dataset, the subjective image quality (Image quality), amount of calcium (Calcium score), average total overlap (OV, in \%), overlap until first error (OF, in \%), clinically relevant overlap (OT, in \%) and inner accuracy (AI, in mm) are listed. In addition, the average over all datasets is shown. All results were obtained with leave-one-image-out cross-validation. Detailed definitions of overlap and accuracy metrics are provided in \citep{Scha09}.}
\label{tab:resultscat08}
\resizebox{\linewidth}{!}{
\begin{tabular}{rllrrrr}
Dataset & Image quality & Calcium score & OV & OF & OT & AI \\ \hline
0 & Moderate & Moderate & 96.6 & 78.2 & 97.9 & 0.30 \\
1 & Moderate & Moderate & 97.9 & 99.5 & 99.2 & 0.24 \\
2 & Good & Low & 98.2 & 100.0 & 100.0 & 0.20 \\
3 & Poor & Moderate & 86.3 & 68.5 & 87.4 & 0.24 \\
4 & Moderate & Low & 97.6 & 95.9 & 98.4 & 0.17 \\
5 & Poor & Moderate & 97.9 & 97.7 & 98.3 & 0.24 \\
6 & Good & Low & 98.7 & 100.0 & 100.0 & 0.18 \\
7 & Good & Severe & 92.7 & 56.9 & 95.4 & 0.23 \\ \hline 
Average & & & 95.7 & 87.1 & 97.1 & 0.23 \\
\end{tabular}}
\end{table}

Table \ref{tab:resultscat08} lists average overlap and accuracy results for each of the eight datasets in the training set. 
In one vessel (Dataset 0, Vessel 0), tracking from a single seed point turned out to be insufficient and an additional point was manually placed.
In terms of overlap, our method obtained an average OV of 95.7\%, an average OF of 87.1\%, and an average OT of 97.1\%. In terms of accuracy, our method obtained an AI of 0.23 mm, which is smaller than the typical width of a voxel in the dataset. In comparison, the average interobserver difference for all points in the training set is 0.20 mm.
Reduced centerline overlap was caused by either undersegmentation or oversegmentation. In one case, coronary artery bridging caused premature termination of the tracker. 
In several cases, the tracker partially followed a connected or adjacent vein or artery segment after reaching the artery's most distal point. 

Performance of the method was also evaluated on the 24 CCTA images in the test set of CAT08, requiring extraction of 96 coronary arteries. For this, we retrained the CNN using the eight CAT08 training scans. Extracted centerlines in the test set were submitted to the CAT08 challenge and an independent evaluation was performed (Table \ref{tab:resultscat08test}). All vessel centerlines were extracted using a single seed point, except for two vessels in Dataset 26 that were hampered by substantial motion artifacts. For both of these vessels, an additional seed point was required. Overlap values were slightly lower than in the training set, with an OV of 93.7\%, an OF of 81.5\%, and an OT of 97.0\%. The estimated location of centerline points was more accurate than in the training set, with an AI of 0.21 mm. Overlap and accuracy results correspond to an average rank of 5.76 in the CAT08 framework, which places the method in third place behind the interactive methods described in \citep{Frim10} and \citep{Scha11} that require both the start- and end-point of the coronary artery, and above 22 other methods for which results are publicly available\footnote{\url{http://coronary.bigr.nl/preview/NWF56B}}. 

\begin{table}[tp]
\centering
\caption{Results for centerline extraction in the CAT08 \textit{test} set. For each dataset, the subjective image quality (Image quality), amount of calcium (Calcium score), average total overlap (OV, in \%), overlap until first error (OF, in \%), clinically relevant overlap (OT, in \%) and inner accuracy (AI, in mm) are listed. In addition, the average over all datasets is shown. Detailed definitions of overlap and accuracy metrics are provided in \cite{Scha09}.}
\label{tab:resultscat08test}
\resizebox{\linewidth}{!}{
\begin{tabular}{rllrrrr}
Dataset & Image quality & Calcium score & OV & OF & OT & AI \\ \hline
8 & Poor & Low & 89.0 & 73.0 & 92.6 & 0.27 \\
9 & Good & Low & 95.6 & 94.3 & 99.2 & 0.18 \\
10 & Moderate & Moderate & 92.9 & 95.4 & 99.1 & 0.22 \\
11 & Good & Moderate & 92.7 & 54.2 & 93.8 & 0.26 \\
12 & Good & Moderate & 92.2 & 50.3 & 95.7 & 0.23 \\
13 & Moderate & Low & 97.7 & 99.8 & 100.0 & 0.16 \\
14 & Moderate & Severe & 95.1 & 84.8 & 99.2 & 0.23 \\
15 & Moderate & Moderate & 94.5 & 95.5 & 98.5 & 0.19 \\
16 & Good & Low & 95.7 & 99.8 & 97.4 & 0.19 \\
17 & Poor & Severe & 83.5 & 41.3 & 85.3 & 0.28 \\ 
18 & Good & Moderate & 94.2 & 79.8 & 98.4 & 0.19 \\
19 & Moderate & Moderate & 93.5 & 100.0 & 100.0 & 0.21 \\
20 & Moderate & Moderate & 98.1 & 93.6 & 99.8 & 0.23 \\
21 & Good & Low & 97.5 & 99.1 & 99.1 & 0.16 \\
22 & Good & Low & 96.2 & 100.0 & 100.0 & 0.20 \\
23 & Moderate & Moderate & 95.5 & 99.6 & 99.7 & 0.20 \\
24 & Moderate & Severe & 95.8 & 77.4 & 98.7 & 0.16 \\
25 & Good & Moderate & 93.8 & 70.4 & 96.7 & 0.23 \\
26 & Poor & Low & 75.7 & 14.7 & 81.1 & 0.31 \\
27 & Good & Moderate & 90.4 & 66.9 & 95.7 & 0.24 \\
28 & Good & Low & 96.1 & 94.4 & 99.2 & 0.15 \\
29 & Poor & Moderate & 98.6 & 77.9 & 99.2 & 0.18 \\
30 & Good & Low & 96.4 & 95.2 & 99.4 & 0.17 \\
31 & Good & Moderate & 97.1 & 98.2 & 99.2 & 0.15 \\ \hline
Average & & & 93.7 & 81.5 & 97.0 & 0.21 \\
\end{tabular}}
\end{table}

We found that subjective image quality as observed by the CAT08 challenge organizers correlated with the performance of our method. In test images that were rated as 'poor' our method obtained an average OV of 86.7\% and an AI of 0.26 mm, while in images rated as 'good' the method obtained an average OV of 94.8\% and an AI of 0.20 mm. The presence of calcium had a less pronounced effect, with an average OV of 91.5\% and average AI of 0.22 in patients with severe calcification and an average OV of 93.3\% and average AI of 0.20 mm in patients with low calcium scores.

\begin{figure}
\centering
\includegraphics[width=0.7\linewidth]{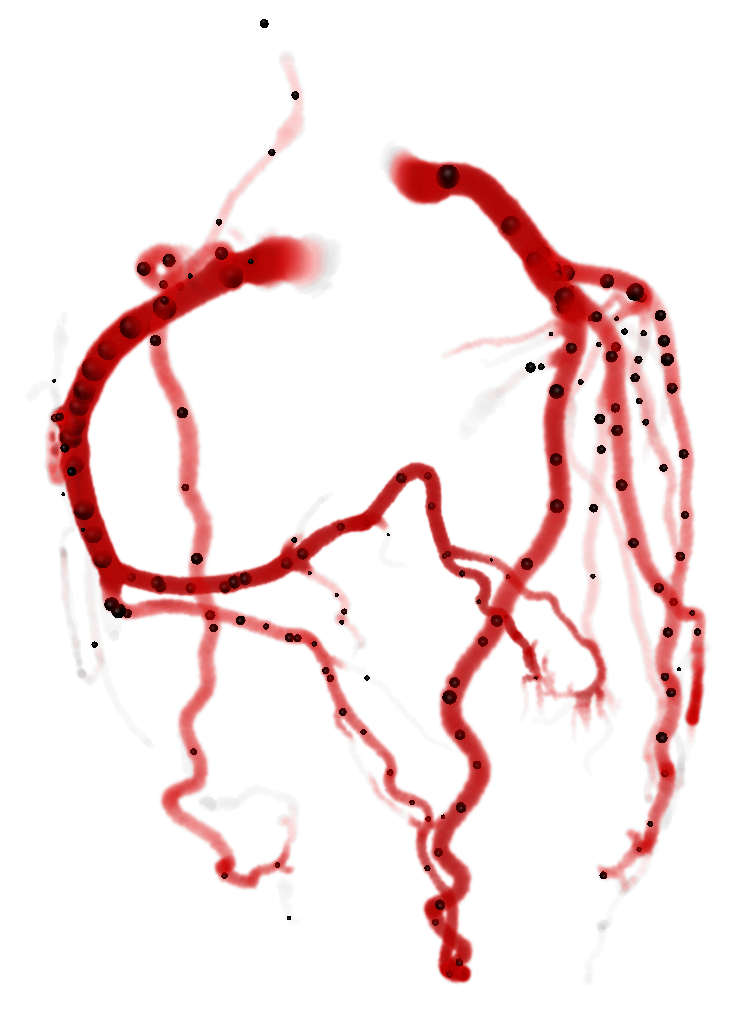}
\caption{Volume rendering of 177 extracted centerlines with estimated radii in a patient from the UMCU set. Black spheres indicate manual reference markers with radii set by an expert. Each marker was used once as a seed point to extract a centerline. The opacity at each location indicates the number of centerlines in the vicinity, indicating that most centerlines were tracked back to the coronary ostia.}
\label{fig:examplewithseeds}
\end{figure}

\subsubsection{Centerline extraction robustness} 
\label{sec:resultsumcu}
The trained CNN was evaluated using 50 additional CCTA images acquired on a Philips CT scanner in our institution, containing 5,448 manually placed markers. 
Each of these markers was used as a seed point, in order to extract 5,448 coronary artery centerlines. 
Fig. \ref{fig:examplewithseeds} shows an example of markers placed by the human expert and all vessels extracted using those markers as seed points. The vessel opacity at each point in the rendering corresponds to the number of centerlines near that point. Hence, most centerlines pass through the proximal coronary arteries, and fewer centerlines pass through coronary artery branches.

\begin{figure}
\centering
\includegraphics[width=\linewidth]{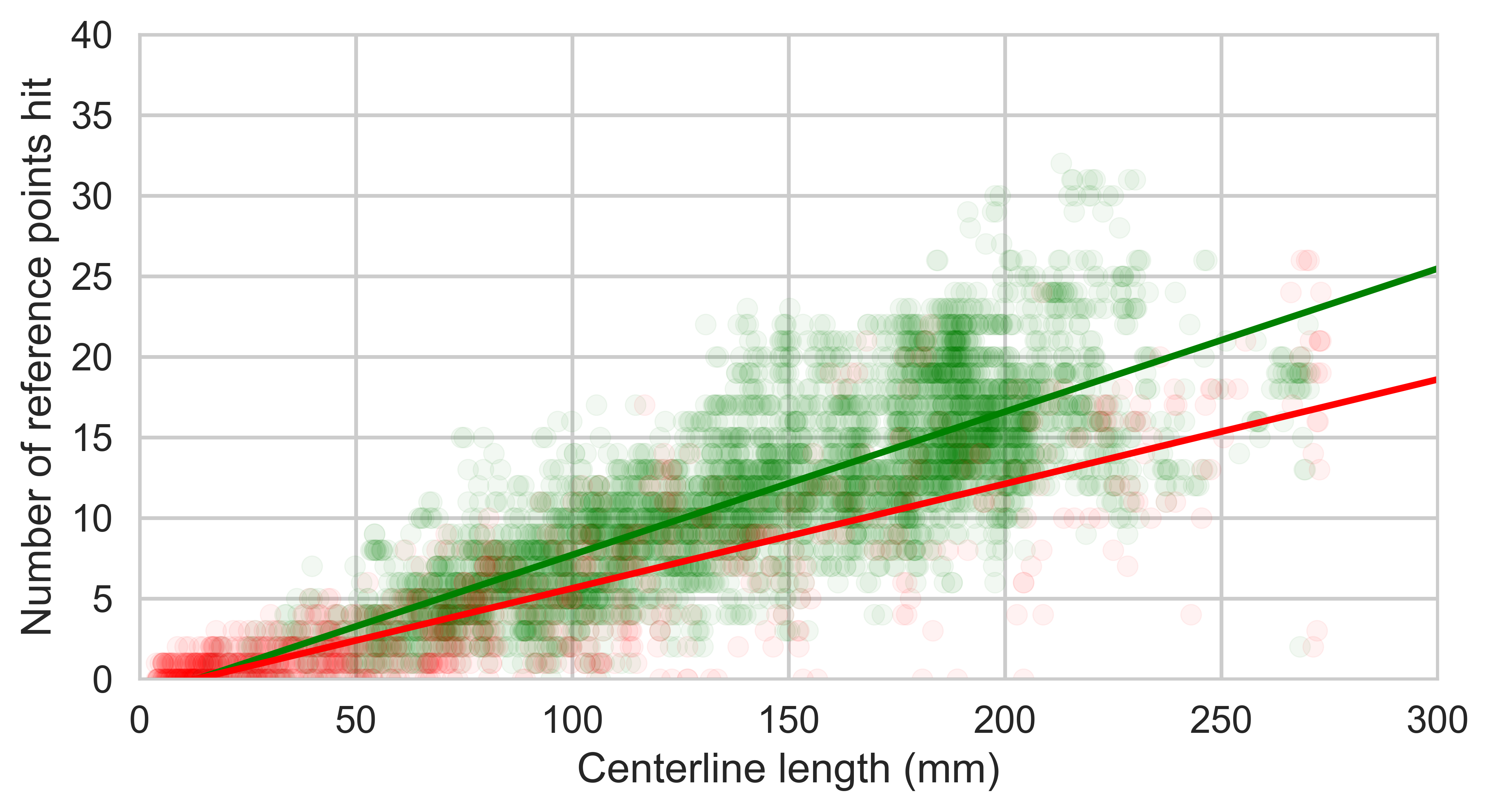}
\caption{Number of reference markers hit as a function of the length of the artery in 5,448 centerlines. Each point corresponds to a centerline, green and red markers indicate whether the centerline reached the coronary ostium or not. The solid lines show linear fits for both groups of centerlines.}
\label{fig:metric}
\end{figure}

For each extracted centerline, we quantified the agreement with manually placed markers. Unlike in the CAT08 set, we did not have point-wise annotated reference centerlines. Hence, agreement was not computed using overlap and accuracy metrics, but in terms of hits. 
A hit was defined as a manually placed marker being within radius distance of an extracted centerline point. In addition, we evaluated whether extracted centerlines reached the manually annotated coronary ostium. 
Fig. \ref{fig:metric} shows a scatter plot in which each data point corresponds to an extracted centerline, and marker colors indicate whether the ostium was reached.
The plot indicates a linear relation between centerline length and number of hits with manually annotated centerline points. Because markers were placed approximately equidistantly at 10 mm intervals, this means that in most cases extracted centerlines followed these markers. The slope of the fitted line for centerlines that did not reach the ostium is smaller, indicating centerline extraction along non-coronary vessels, or on the far left side, indicating short centerlines that failed to reach the ostium. Short centerlines may have been due to artifacts being encountered, but also to erroneous seeding in coronary veins.

\begin{figure}
\centering
\subfloat[]{
\includegraphics[width=0.29\linewidth]{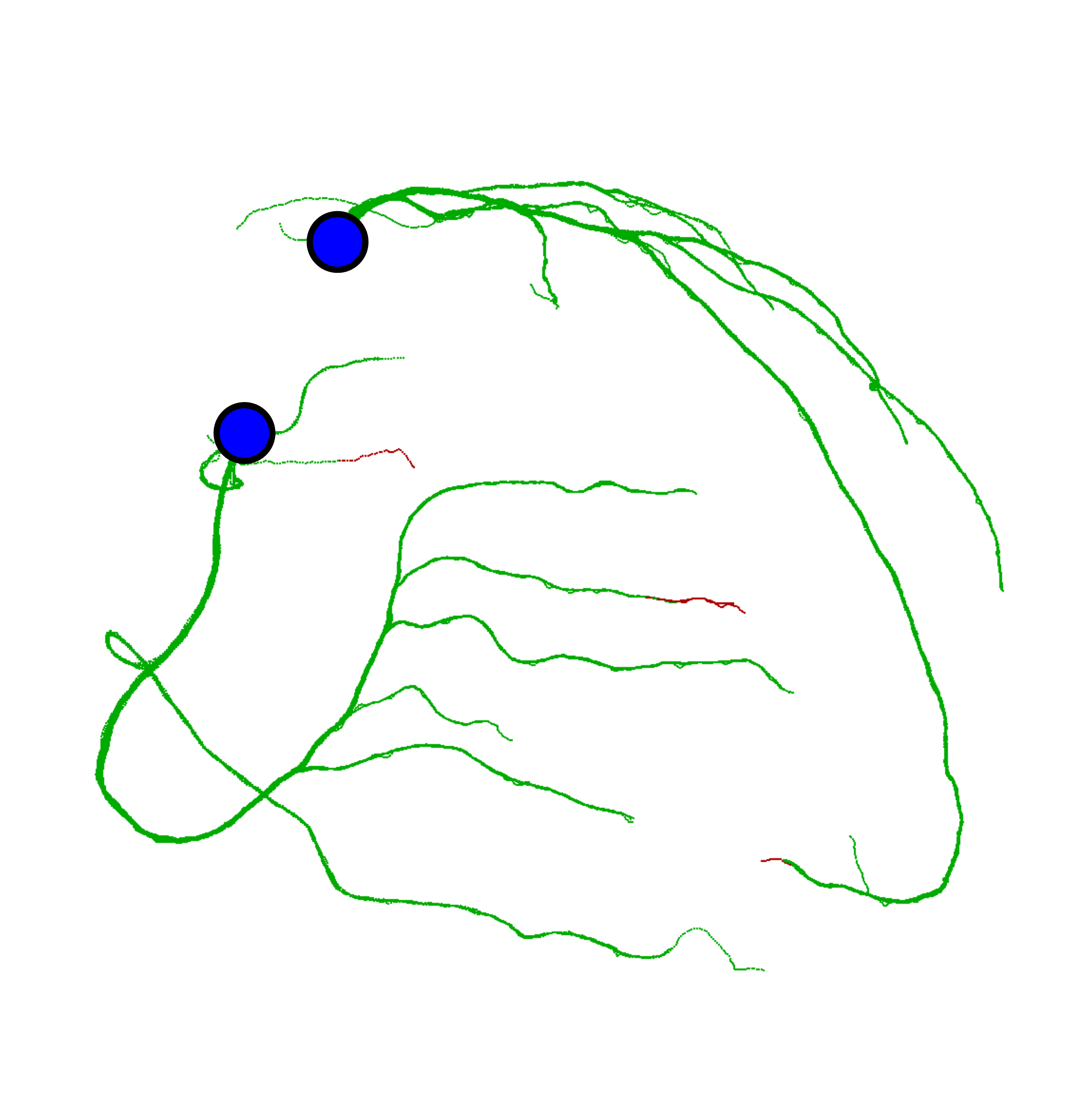}
\label{subfig:error2464}
}
\subfloat[]{
\includegraphics[width=0.29\linewidth]{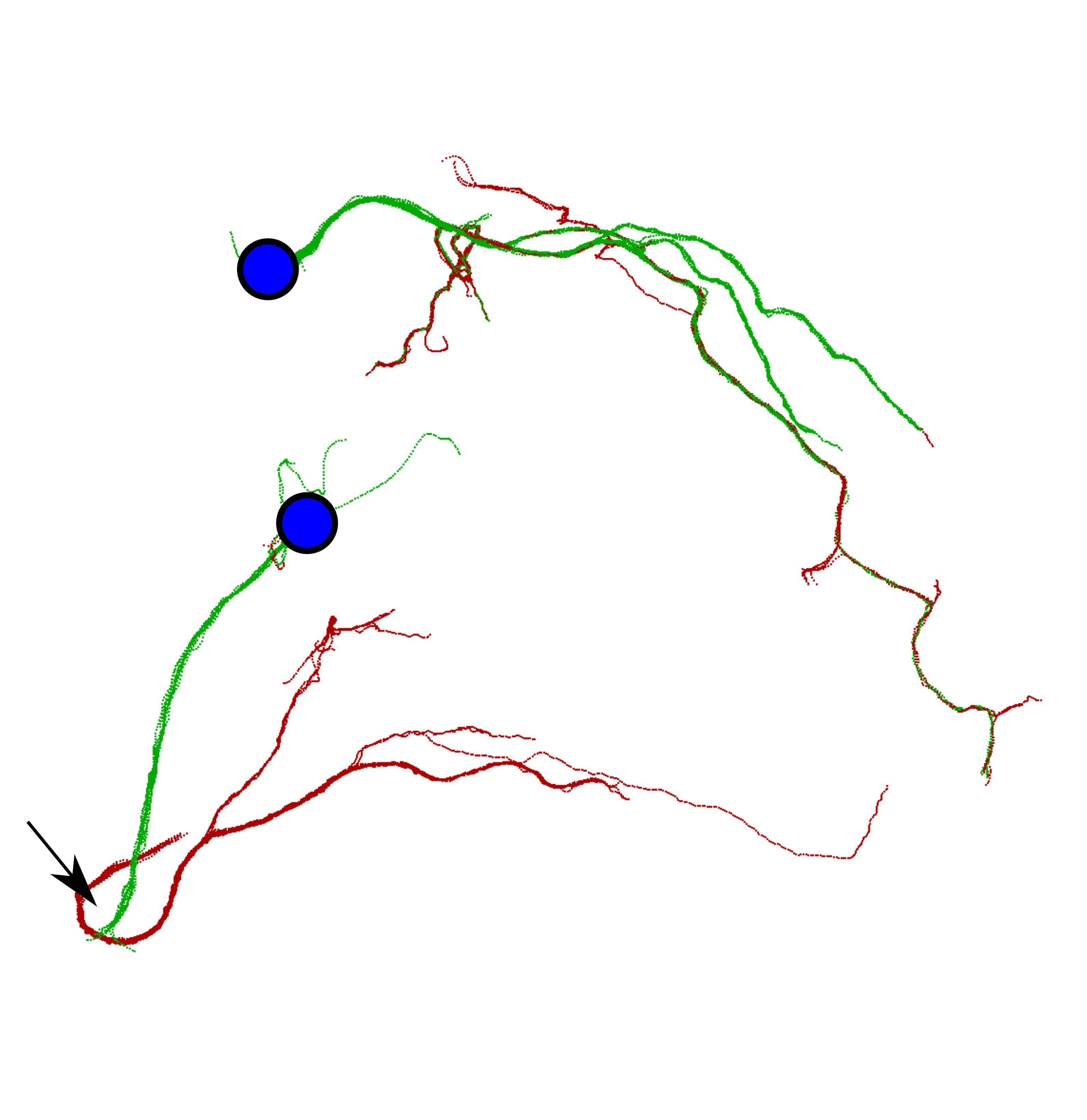}
\label{subfig:error2068}
} 
\subfloat[]{
\includegraphics[width=0.29\linewidth]{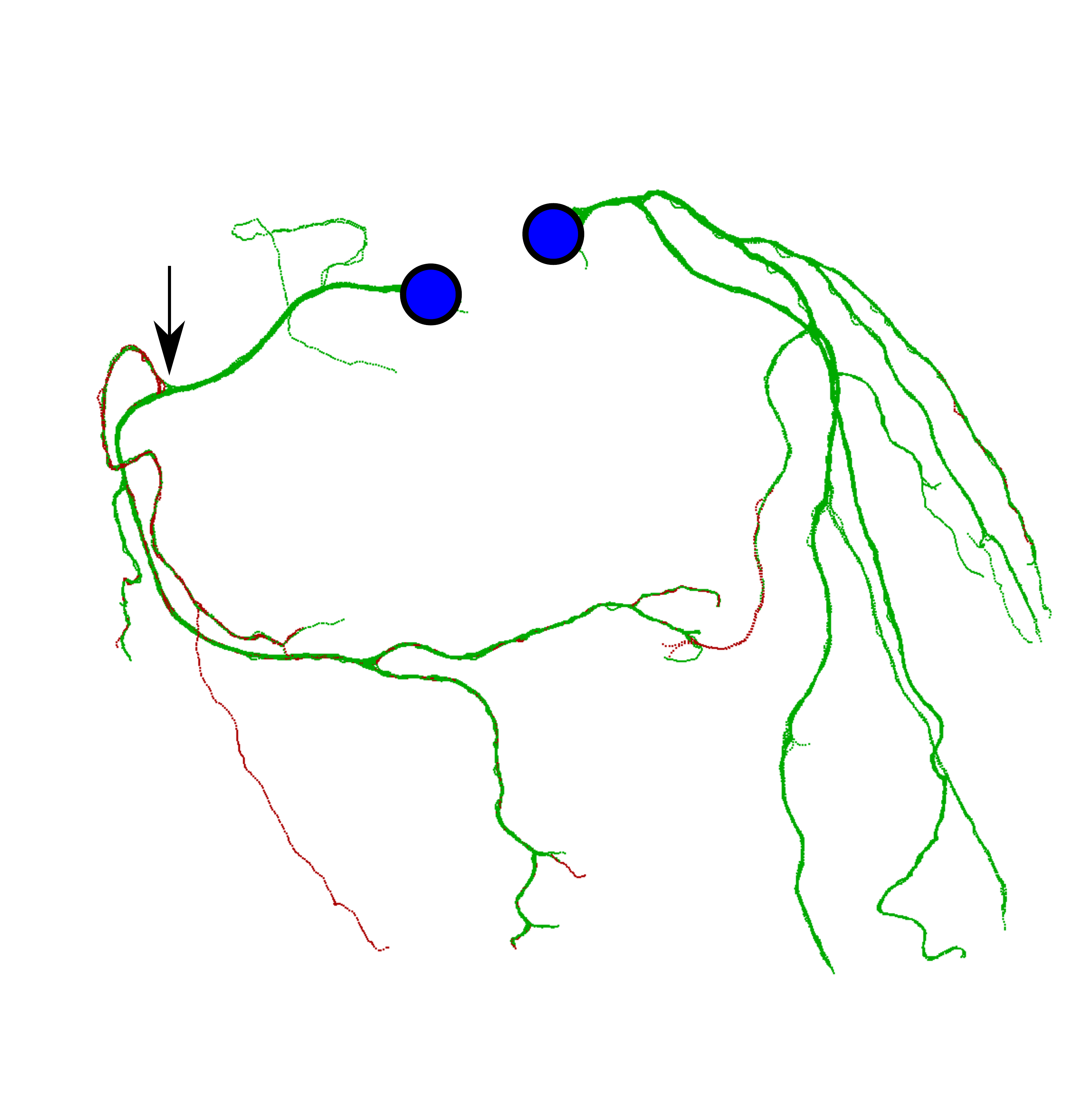}
\label{subfig:error10008}
}
\caption{Centerline extraction in a UMCU image. Green and red indicate whether the centerline reached the coronary ostium or not. The blue spheres indicate the location of the ostia. In \protect\subref{subfig:error2464} paths are found between almost all markers and an ostium. In \protect\subref{subfig:error2068} a large step-and-shoot artifact in the RCA (black arrow) prevents most centerlines seeded in the distal RCA from reaching the right coronary ostium. In \protect\subref{subfig:error10008} several small branches do not reach the ostium. In addition, one larger branch (black arrow) makes a sharp angle with the main branch, leading the tracker away from the ostium, instead of towards it.}
\label{fig:errors}
\end{figure}

\begin{figure*}
\centering
\subfloat[]{
\includegraphics[width=0.48\linewidth]{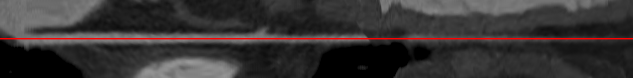}
\label{subfig:intensitydrop} 
} 
\subfloat[]{
\includegraphics[width=0.48\linewidth]{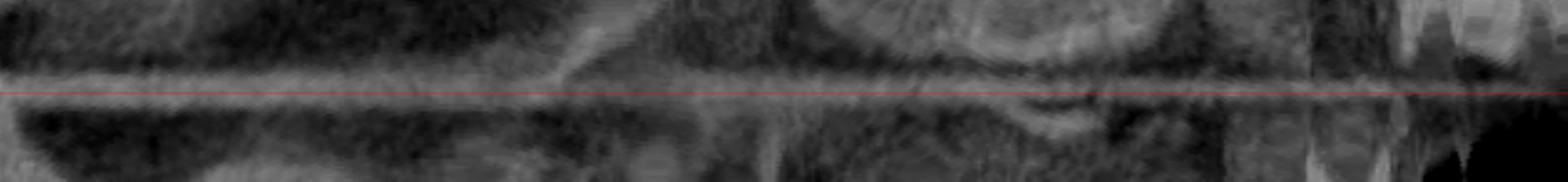}
\label{subfig:motionartifact}
} \\
\subfloat[]{
\includegraphics[width=0.48\linewidth]{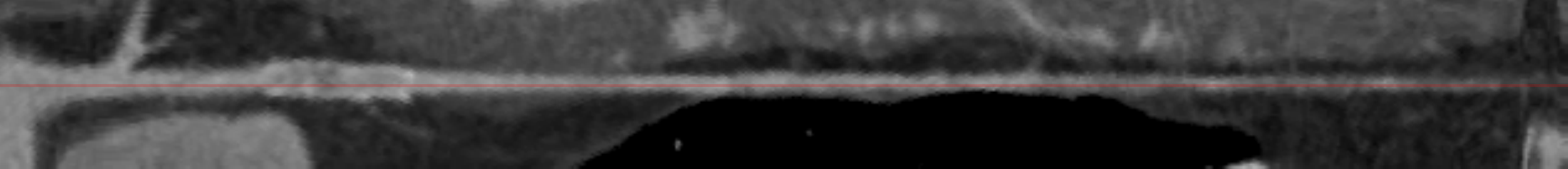}
\label{subfig:properwithstent}
} 
\subfloat[]{
\includegraphics[width=0.48\linewidth]{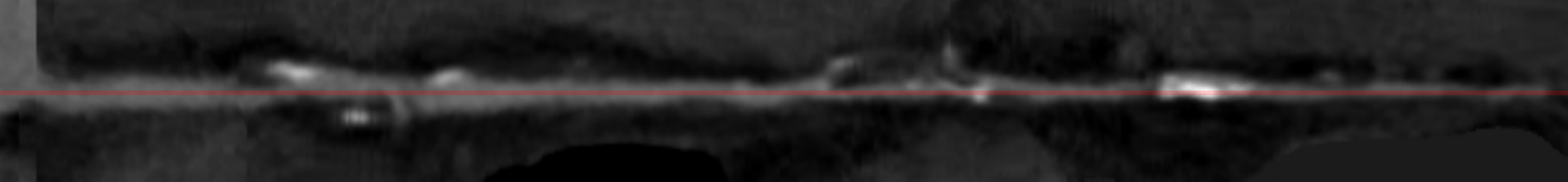}
\label{subfig:calc}
} 
\caption{Stretched multi-planar reconstructions of successfully extracted coronary artery centerlines (in red) in the presence of a \protect\subref{subfig:intensitydrop} step artifact, \protect\subref{subfig:motionartifact} motion artifact, \protect\subref{subfig:properwithstent} intravascular stent, and \protect\subref{subfig:calc} coronary calcification.}
\label{fig:successes}
\end{figure*}

The plot in Fig. \ref{fig:metric} shows overall results on the 50 CCTA images in the UMCU dataset. However, an analysis on individual images reveals some typical errors that the method makes. Centerline extraction was most affected by stepping artifacts introduced by step-and-shoot image acquisition (Fig. \ref{subfig:error2068}). In some cases, centerlines failed to reach the ostium because of anatomical characteristics such as severe calcifications or stenosis, or vessel branching with sharp angles (Fig. \ref{subfig:error10008}). Nevertheless, the method was able to extract the centerline in cases where step artifacts were not very substantial (Fig. \ref{subfig:intensitydrop}), mild motion artifacts were present (Fig. \ref{subfig:motionartifact}), intravascular stents could be identified (Fig. \ref{subfig:properwithstent}) and coronary artery calcification was present in the arteries (Fig. \ref{subfig:calc}). In practice, many limitations may be overcome by placement of additional seed points.

\begin{figure}
\centering
\includegraphics[width=\linewidth]{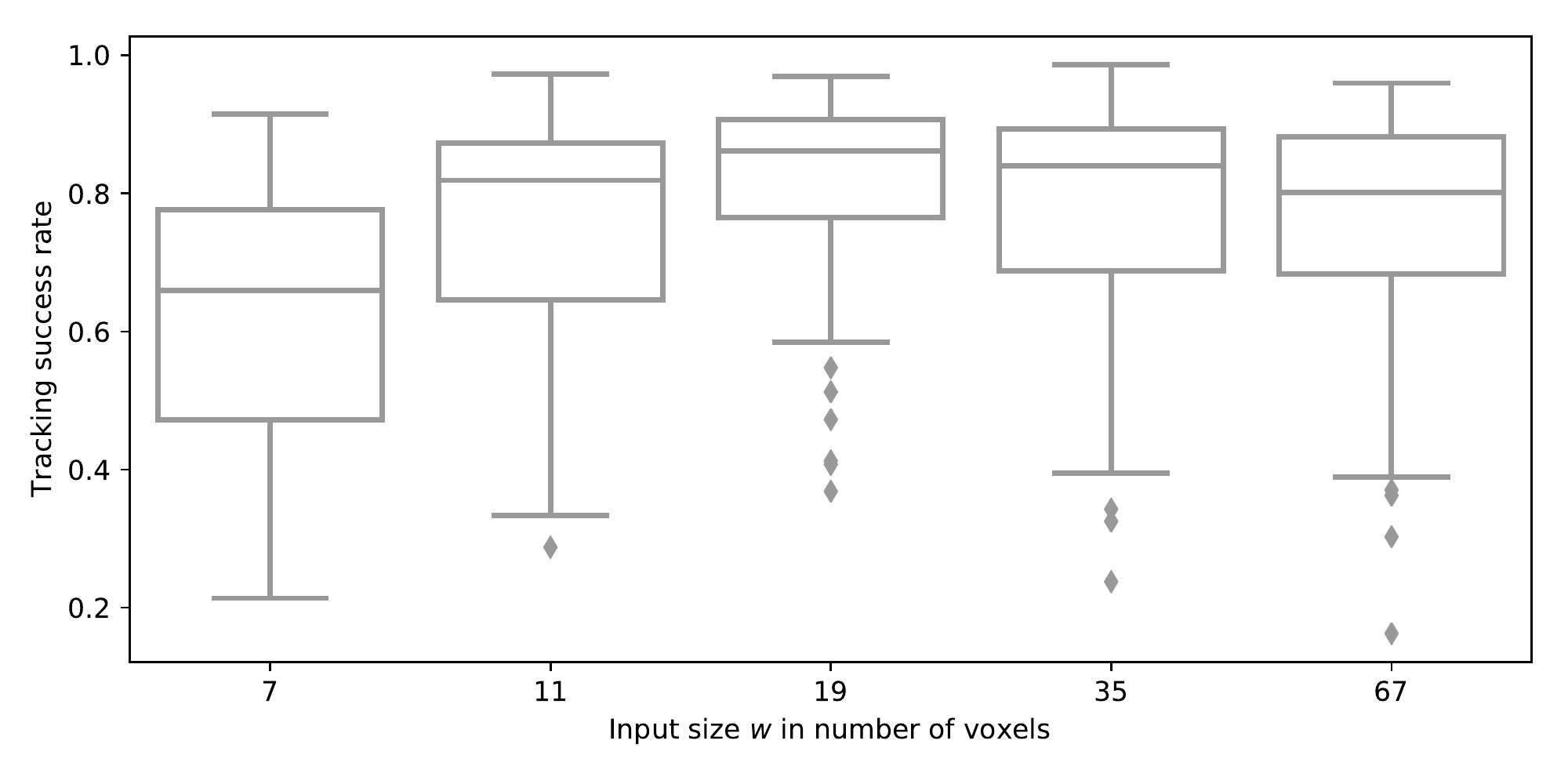}
\caption{Box plots showing the influence of the receptive field size on the success rate of the algorithm in CCTA scans in the UMCU dataset. By changing the input size $w$ of the CNN, the success rate first increases and then decreases. Too small receptive fields may not contain sufficient information for the tracker, while too large receptive fields may lead to overfitting of the CNN.}
\label{fig:networksize}
\end{figure}

\begin{figure}
\centering
\includegraphics[width=\linewidth]{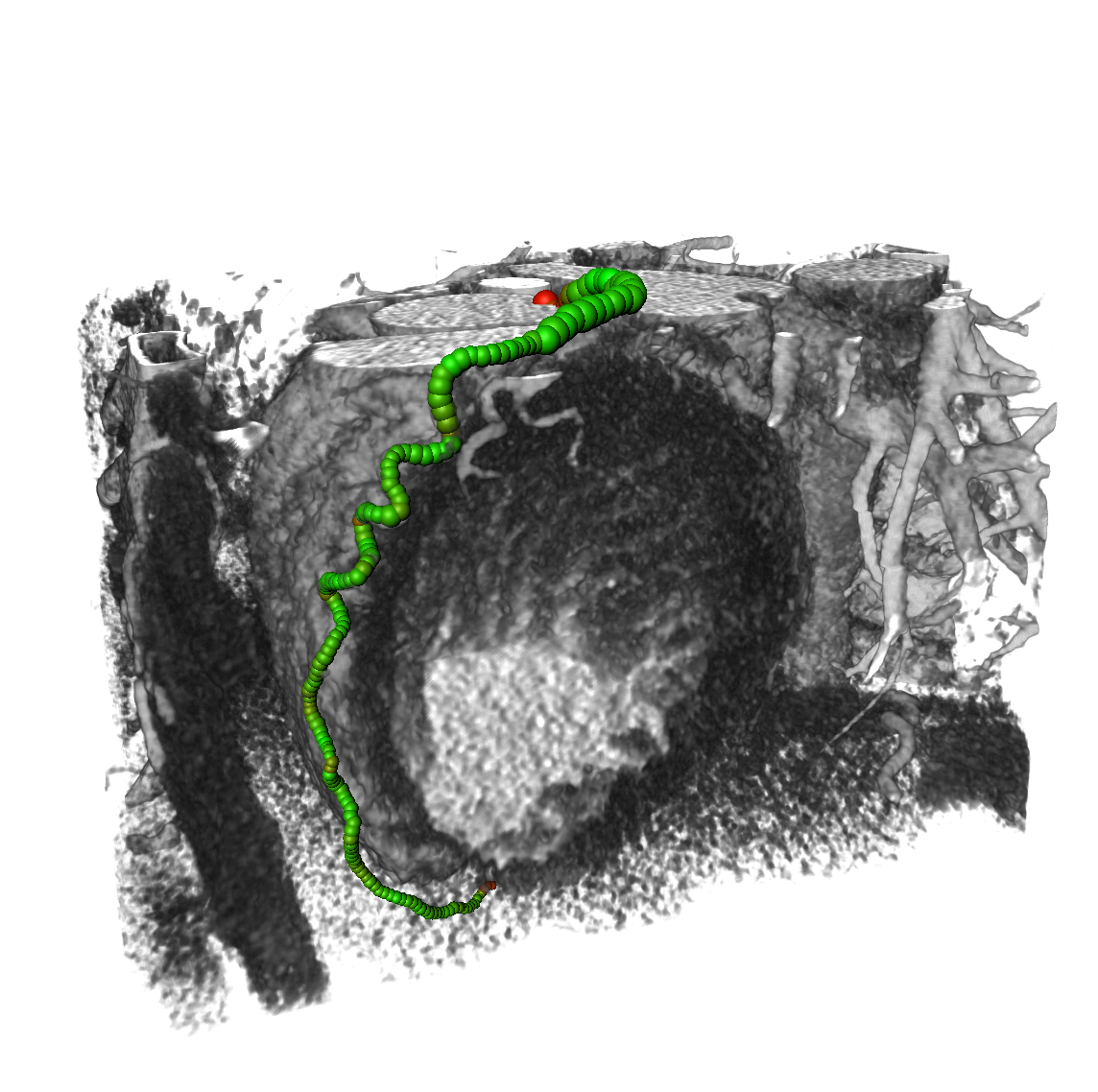} 
\caption{Successful centerline extraction in a patient with tortuous coronary arteries. Spheres correspond to centerline points, their radii to automatically determined radius values, and their colors indicate the uncertainty of the classifier. Green corresponds to low entropy values, and red (at the ostium and end of the centerline) corresponds to high entropy.}
\label{fig:oddones}
\end{figure}

The receptive field of the CNN determines the part of the image that is taken into account to estimate the local orientation and radius. The choice of the receptive field size may affect the CNN's performance. We performed an experiment in which we removed or inserted layers with increasing dilation levels in the architecture in Table \ref{tab:network} to reduce the receptive field size from $w=19$ voxels to 7 or 11 voxels, or increase the receptive field to 35 or 67 voxels. We trained these models using the full CAT08 training set and evaluated the tracking success rate in the UMCU dataset, i.e. the ratio of centerlines that reached the ostia (Fig. \ref{fig:networksize}). This shows that a receptive field of $w=19$ leads to the highest success rate. This receptive field allows the CNN to take sufficient information into account to make a correct prediction, without making too many assumptions about the coronary anatomy. For example, Fig. \ref{fig:oddones} shows a patient with tortuous left coronary arteries. For this patient, the success rate was 87\%, 74\% or 68\% for receptive fields of $19$, $35$ or $67$ voxels, respectively. 

\subsubsection{Fully automatic centerline extraction} 
\label{sec:orcaresults}
The orCaScore dataset was used to assess the performance of fully automatic centerline extraction, i.e. tracking initialized with automatically detected seeds. While the CAT08 images were acquired on Siemens CT scanners and the UMCU images were acquired on Philips CT scanners, these 36 images were acquired on GE and Toshiba scanners. Two additional CNNs were trained: one to detect potential tracker seed locations and one to identify the location of the coronary ostia. The CNN for tracker seed detection was trained using coronary masks derived from the centerlines obtained in the UMCU dataset. For this, we selected only vessels that successfully reached the ostium. The CNN for ostia detection was trained using manual ostia annotations in the UMCU dataset.

\begin{figure*}
\centering
\subfloat[]{
\includegraphics[width=0.27\linewidth]{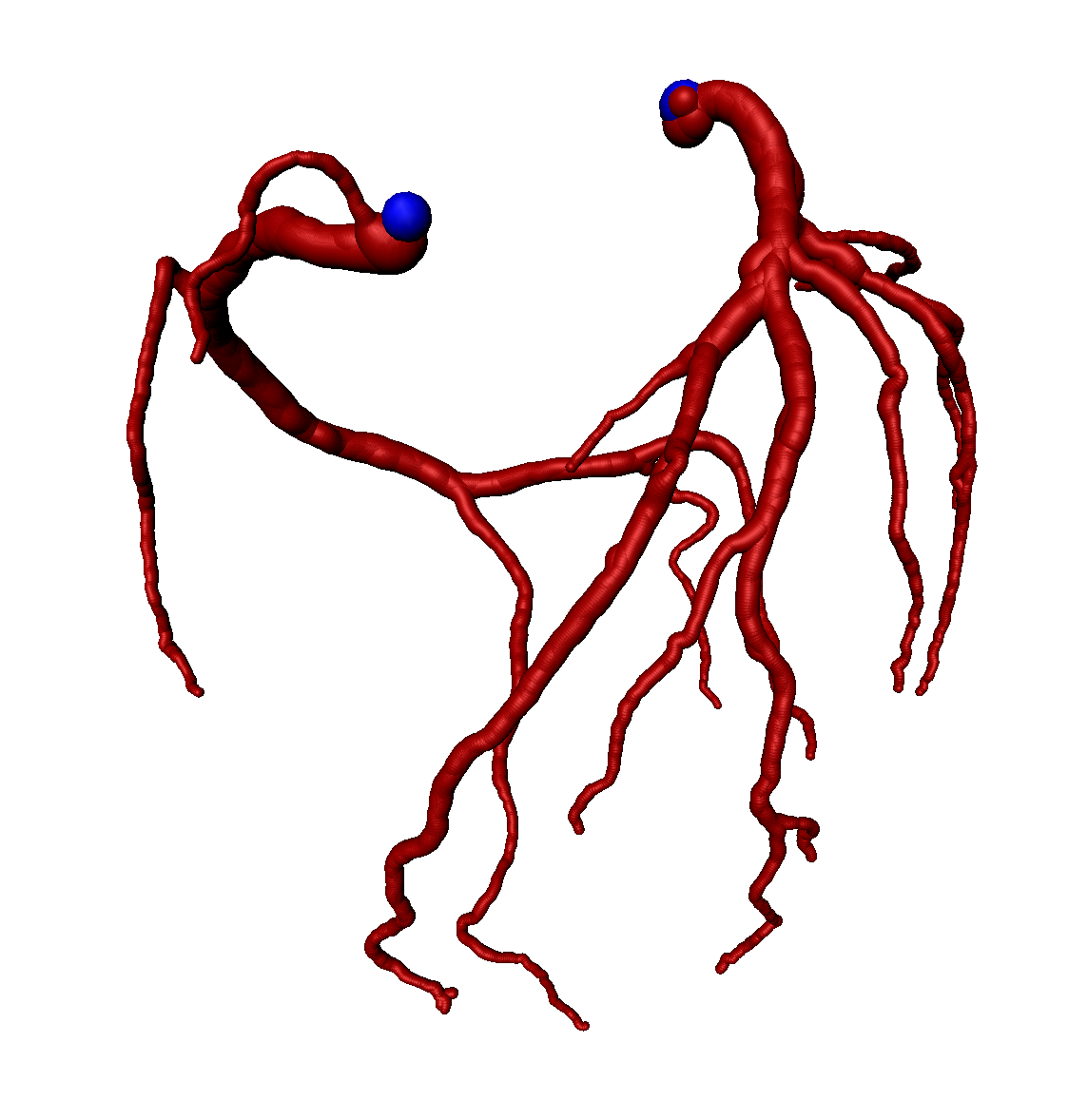}
\label{subfig:fullyautomatica}
}
\subfloat[]{
\includegraphics[width=0.27\linewidth]{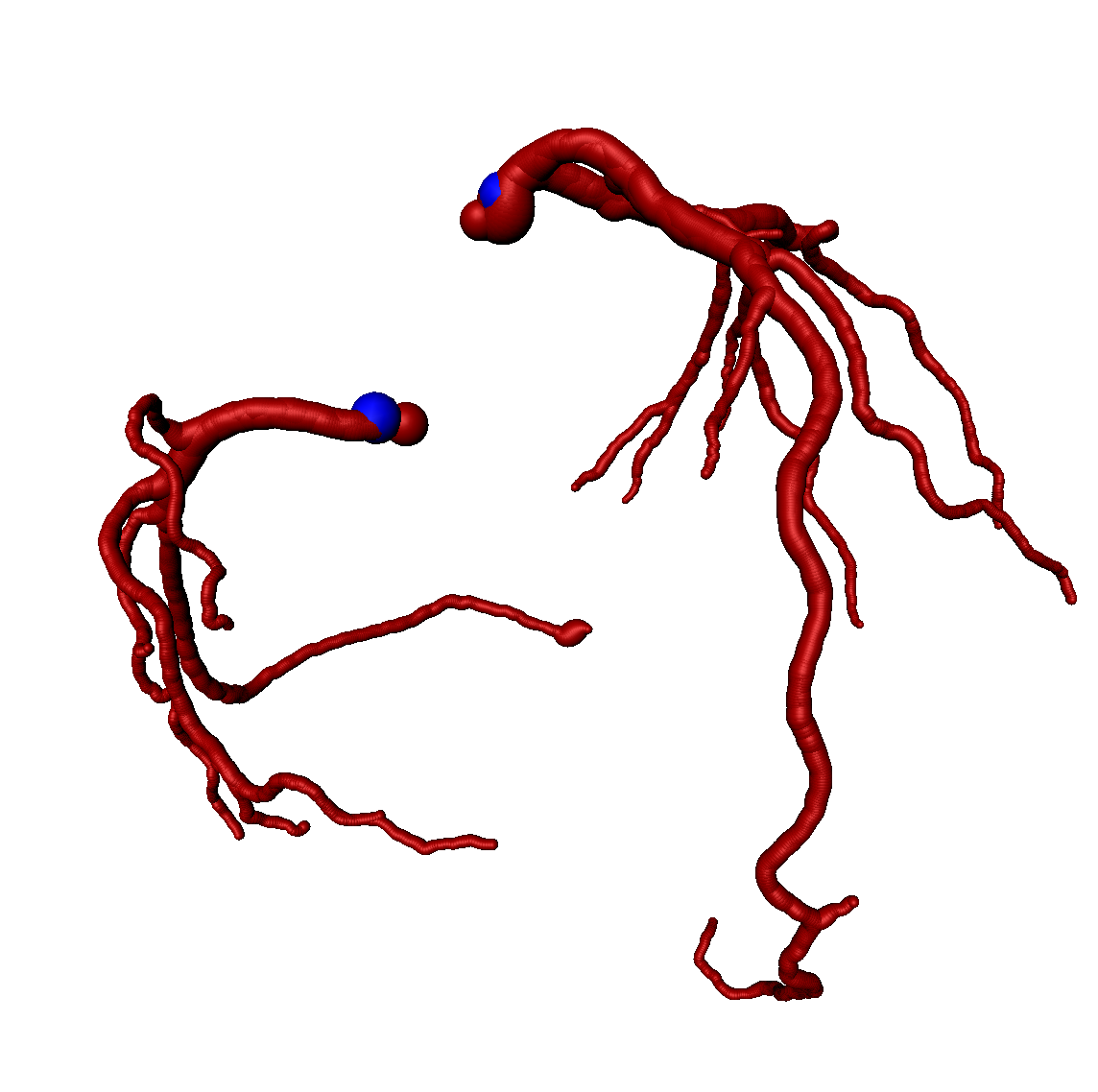}
\label{subfig:fullyautomaticb}
} 
\subfloat[]{
\includegraphics[width=0.27\linewidth]{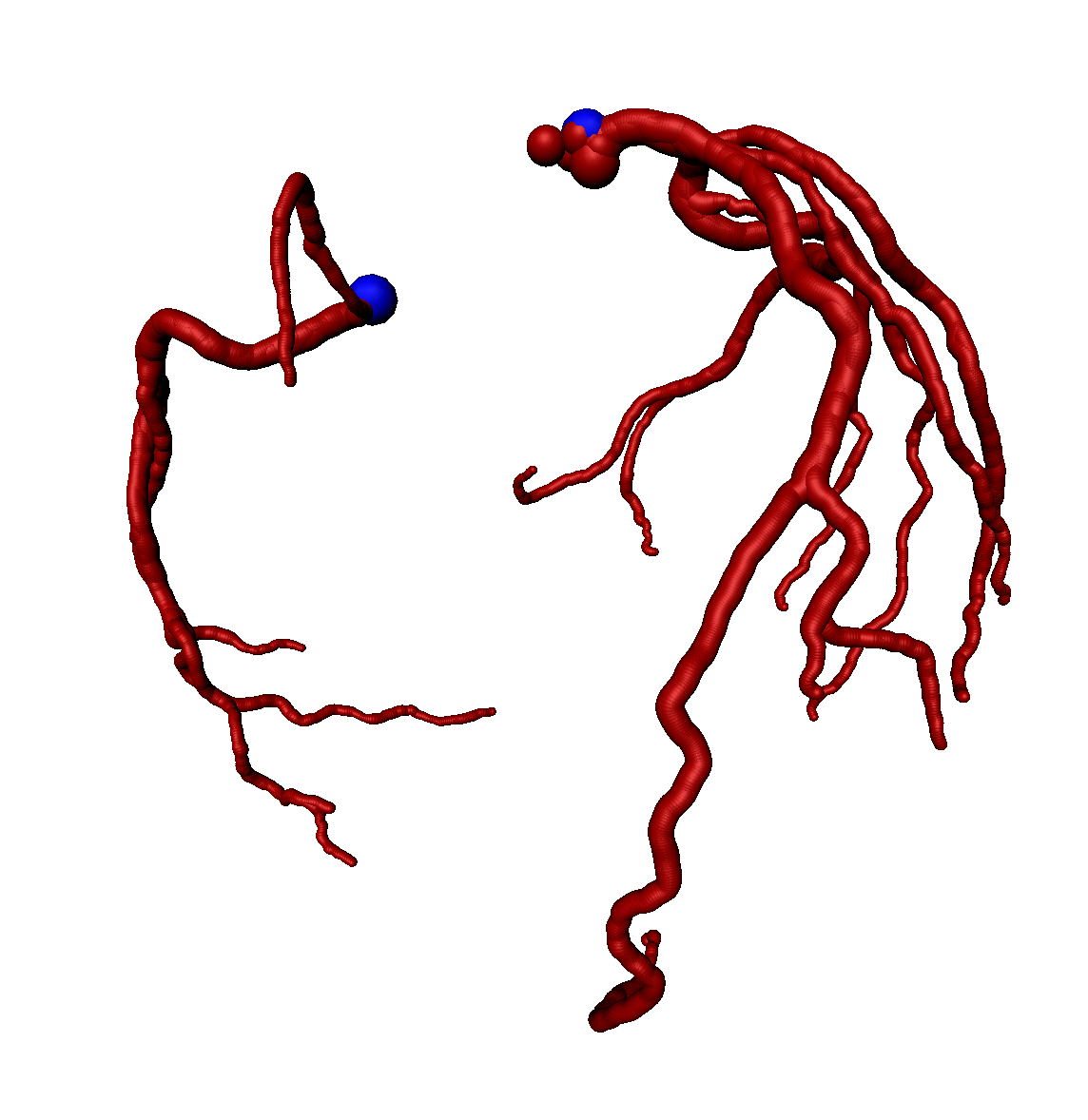}
\label{subfig:fullyautomaticc}
}
\caption{Fully automatic centerline extraction in three CCTA images of the orCaScore dataset acquired on \protect\subref{subfig:fullyautomatica} a GE scanner or \protect\subref{subfig:fullyautomaticb}\protect\subref{subfig:fullyautomaticc} a Toshiba scanner.}
\label{fig:exampleautomatic}
\end{figure*}

For each test scan, 200 seed points for tracker initialization were identified as local maxima in the predicted coronary artery proximity map. Likewise, the two coronary ostia were identified as local maxima in the predicted ostia proximity map. When processing the queue of seed points, around 150 seed points in each scan were typically found to overlap with already extracted centerlines, so that around 50 unique seeds were used for tracker initialization. The centerline extraction CNN was the same as in the previous section and it was trained using the full CAT08 training set.

\begin{figure}
\centering
\includegraphics[width=\linewidth]{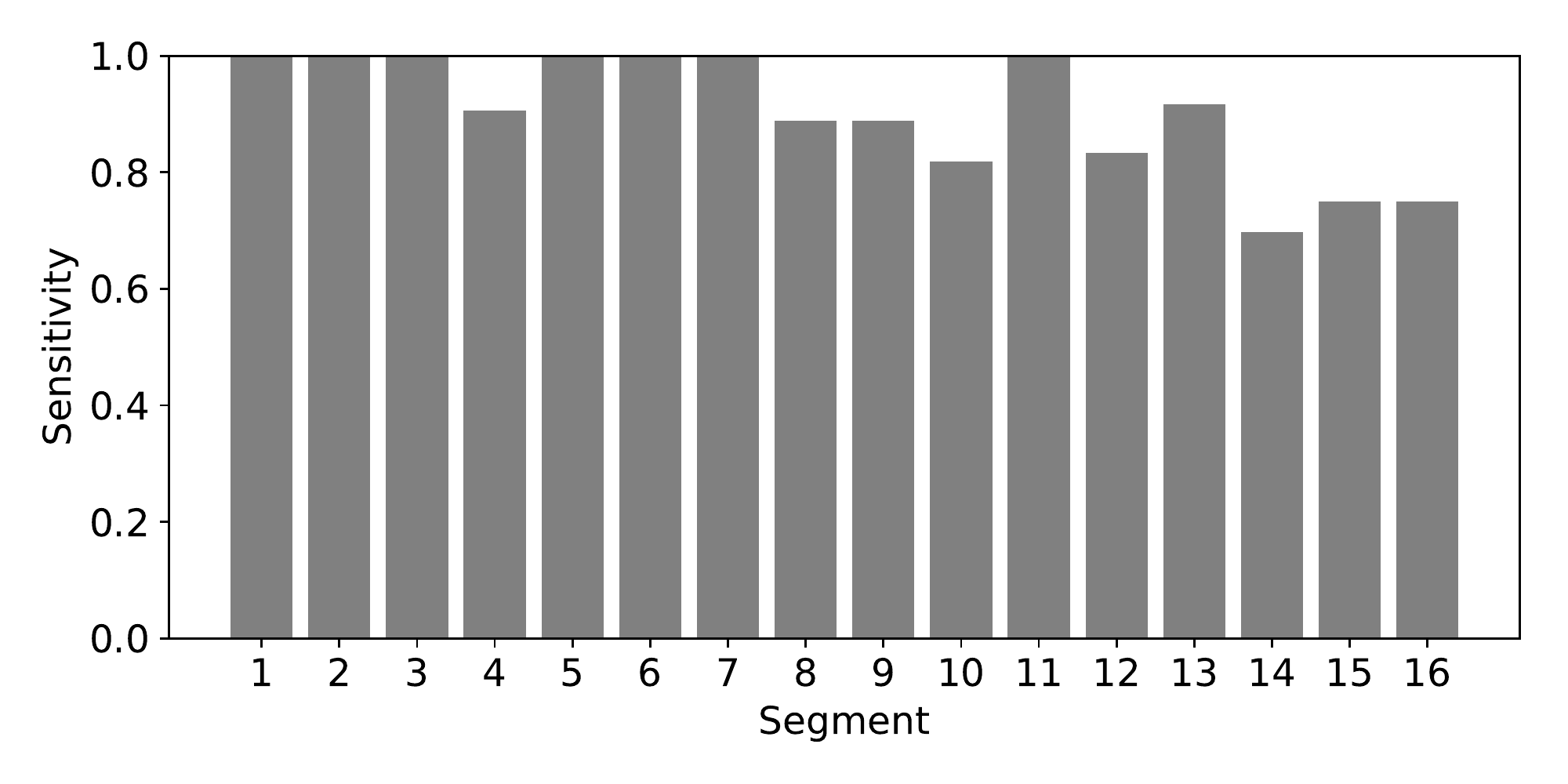}
\caption{Per segment sensitivity across 36 CCTA scans. Segment numbers are: proximal RCA (1), mid RCA (2), distal RCA (3), right posterior descending artery (RPD, 4), left main (LM, 5), proximal LAD (6), mid LAD (7), distal LAD (8), first diagonal branch (9), second diagonal branch (10), proximal LCX (11), first obtuse marginal branch (12), mid LCX (13), first left posterolateral branch (14), left posterior descending branch (15), and ramus intermedius (16). Sensitivity is computed based on those segments visualized in the CCTA image.}
\label{fig:segments}
\end{figure}

We found that automatic ostium detection was successful in all 36 test images, with an average error of $1.8\pm 1.0$ mm with respect to reference annotations. Automatic coronary tree extraction took around 20 s per patient. To assess whether the fully automatic method can successfully extract full coronary artery trees, we performed a qualitative analysis \cite{Zhou12}. For each of 16 clinically relevant coronary artery segments \cite{Hamd11}, an observer scored whether the segment was successfully identified by the method, missed by the method or not visualized in the CCTA scan. Fig. \ref{fig:exampleautomatic} shows examples of fully automatically extracted centerlines. On average, the method was able to identify 92\% of clinically relevant coronary tree segments visualized in a CCTA scan. In 16 out of 36 scans, all segments were identified, and the lowest number of segments identified in a CCTA scan was 10 out of 15. The average number of false negative segments per scan was 1.17.  Fig. \ref{fig:segments} shows the sensitivity of the method for each of these 16 segments across the orCaScore dataset. Whereas the proximal segments of the main coronary branches (segments 1, 6, 11) were identified in all patients, smaller branches such as the left posterolateral branch were occasionally missed. Note that this analysis only includes the 16 coronary segments defined in \cite{Hamd11} and that the method may find additional coronary branches. Similarly, trees may occasionally include false positive responses in veins.
The fully automatic method was also applied to the CAT08 test set and a quantitative evaluation was performed using the CAT08 framework. The four vessel centerlines evaluated in CAT08 were selected from the automatically identified artery trees based on proximity to a uniquely identifying point indicated in each artery by the challenge organizers. We found that quantitative results for overlap and accuracy were similar to those reported in Table \ref{tab:resultscat08test}
, indicating that the fully automatic method could identify the four relevant coronary arteries in each scan. The fully automatic method obtained OV, OF, OT and AI scores of 94.3\%, 79.7\%, 96.9\% and 0.23 mm. These scores are slightly lower than those of the best performing automatic method in CAT08 \cite{Zhen13}.\footnote{\url{http://coronary.bigr.nl/preview/NZ46GA}}.

\begin{figure*}
\centering
\subfloat[CAT08 dataset]{
\includegraphics[width=0.48\linewidth]{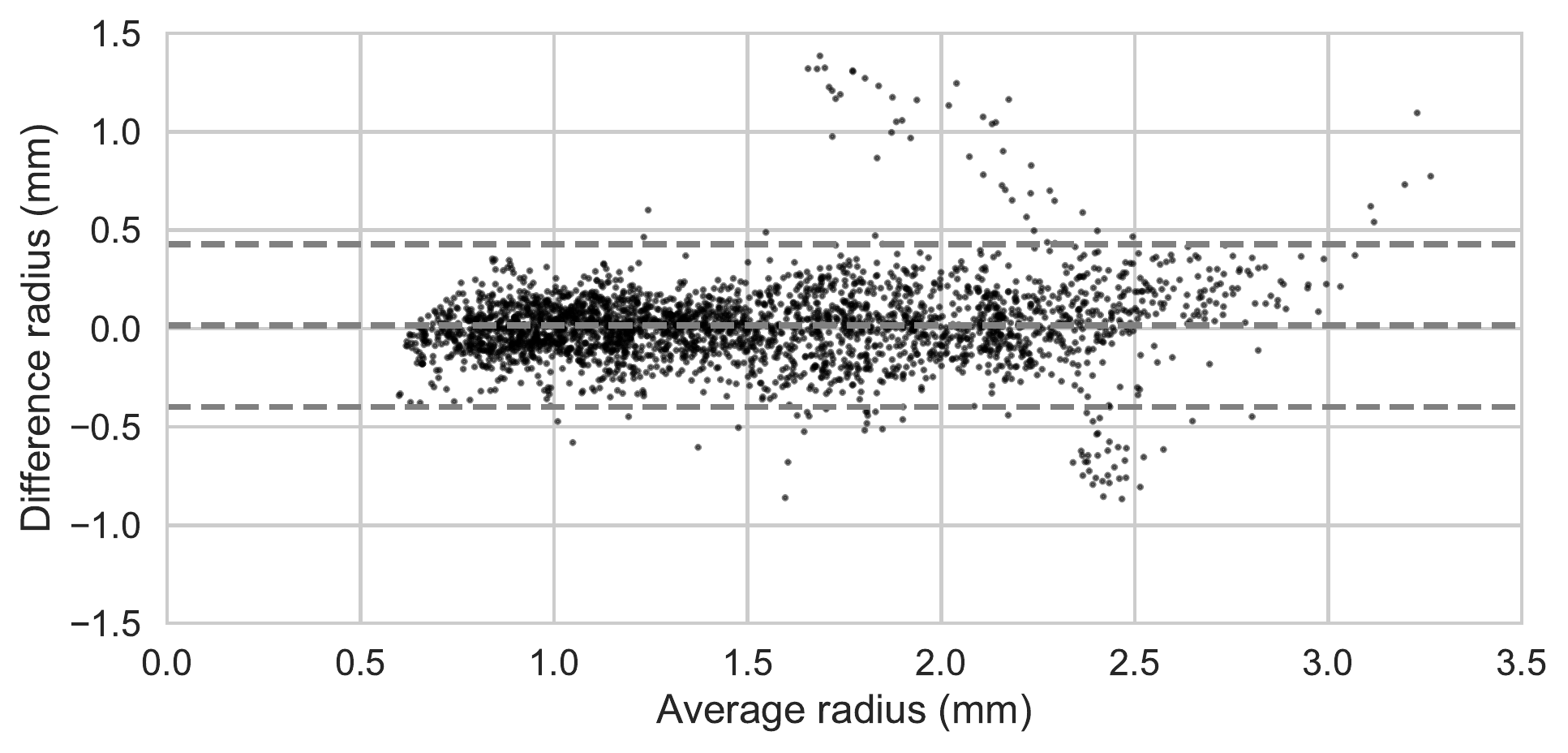}
\label{subfig:bacat}
} 
\subfloat[UMCU dataset]{
\includegraphics[width=0.48\linewidth]{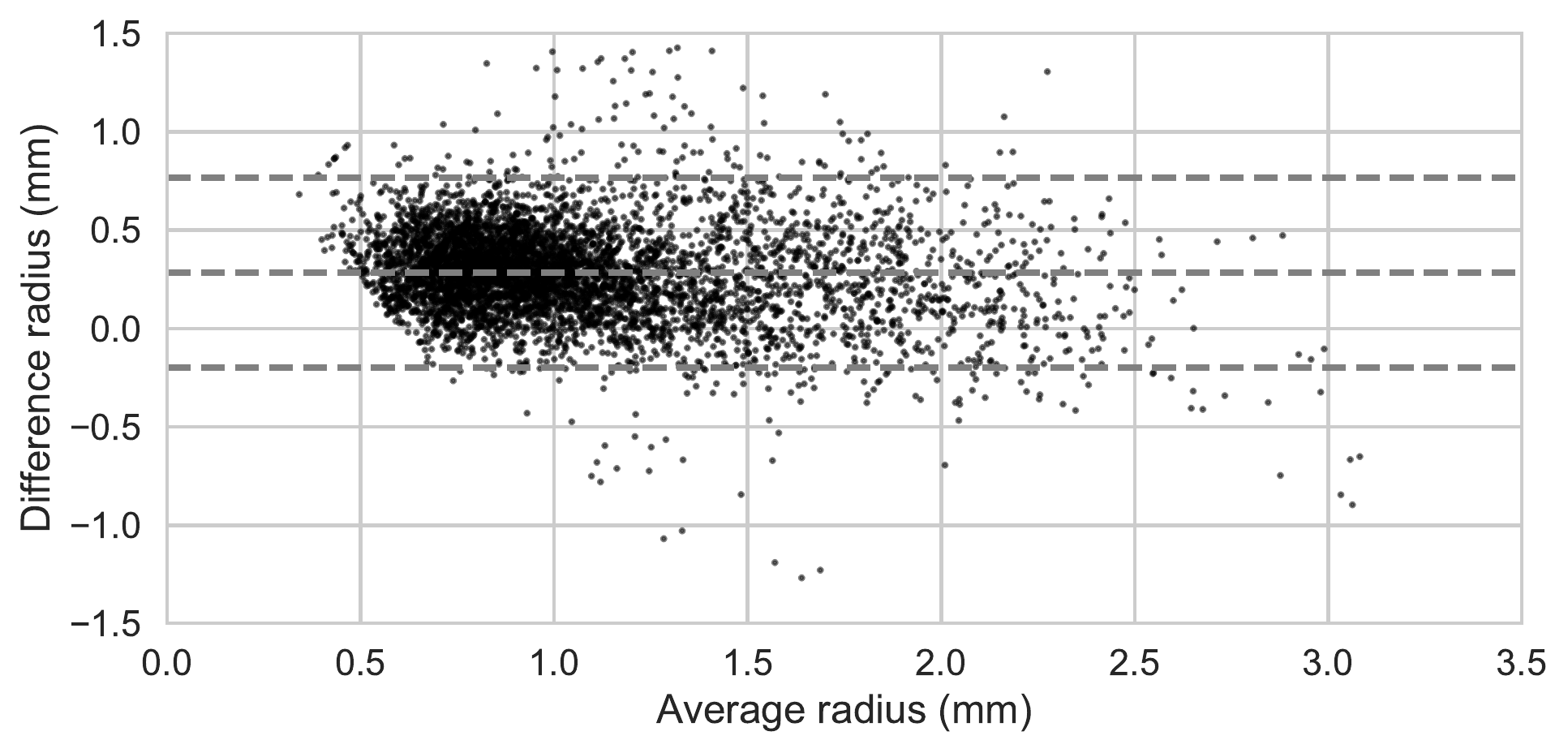}
\label{subfig:baumcu}
}
\caption{Bland-Altman plot comparing reference and automatically determined radius values. The $x$-axis shows the average of the automatically and manually determined radius values (in mm), the $y$-axis shows the difference between the automatically determined and manually determined radius values (in mm). \protect\subref{subfig:bacat} 139,337 centerline points along 32 coronary arteries in the CAT08 dataset. For visualization purposes, every 50th point is shown. Values were obtained using leave-one-image-out cross-validation. \protect\subref{subfig:baumcu} 5,448 centerline points in the UMCU dataset.}
\label{fig:blandaltman}
\end{figure*}

\subsection{Radius estimation}
The regressor output node of the CNN estimates the vessel radius at a location $\mathbf{x}$ in the coronary artery.  To assess how well these radius estimates correspond to reference radius values, we compared automatically determined radius values and manual reference radius values. 

For the CAT08 dataset, we only had access to reference radius values in the eight training images. Hence, we used the models trained during cross-validation (Section \ref{sec:cat08centerline}) to compute the estimated radius value at every reference centerline point in the training images. The Bland-Altman plot in Fig. \ref{subfig:bacat} shows the agreement between reference radii and radii determined by our method, for all training images. There was a negligible systematic underestimation of -0.01 mm. The 95\% limits of agreement were $-0.39$ mm and $0.41$ mm, which shows that for most centerline points radius estimation errors were within the width of a typical CCTA image voxel. 

In addition to the CAT08 scans, we evaluated radius prediction performance at every point in the UMCU scans indicated by the observer (Fig. \ref{subfig:baumcu}). 
While the limits of agreement were similar to the CAT08 dataset, there was a systematic overestimation of the radius (0.31 mm) by the automatic method compared to the manually annotated reference. This difference is partially caused by differences between annotation protocols in the CAT08 and UMCU datasets. The manual radius values in the UMCU dataset were determined in the axial plane of the CCTA image, while in the CAT08 dataset they were determined in cross-sectional reconstructions \citep{Scha09}. Consequently, a difference in radius values is also observed when comparing the distribution of reference radius values in the CAT08 dataset (median [IQR] 1.37 [1.03--1.88] mm) with the distribution of reference radius values in the UMCU dataset (median [IQR] 0.81 [0.61--1.16] mm). Based on this, the CNN may have learned to estimate radius values differently than the expert. To test whether the difference was really caused by the annotation protocol and not by the dataset, the observer used the same protocol to also annotate the coronary arteries in the CAT08 dataset. This resulted in radius values that were closer to reference annotations in the UMCU dataset than those in the CAT08 dataset (median [IQR] 0.87 [0.64--1.32] mm). This is also visible in Fig. \ref{fig:examplewithseeds}, where the extracted vessels have slightly larger radii than the manually placed markers.

\begin{figure}
\centering
\includegraphics[width=\linewidth]{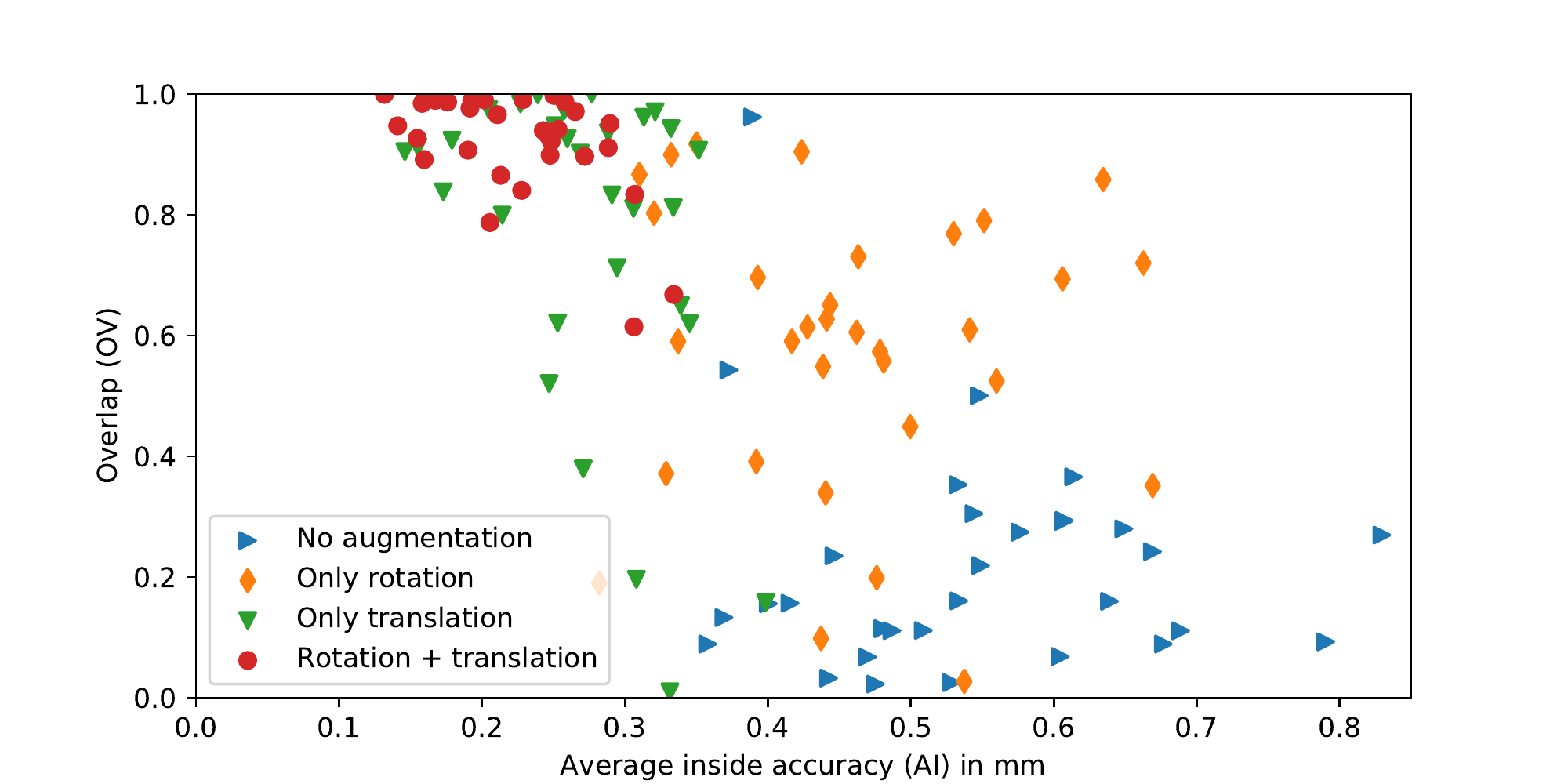}
\caption{The average inside accuracy (AI, lower values are better) and overlap (OV, higher values are better) of 32 automatically extracted centerlines in the CAT08 training set, obtained using different combinations of training data augmentation.}
\label{fig:quantaugment}
\end{figure}

\begin{figure}[tp]
\centering
\subfloat[Trained without off-centerline points]{
\includegraphics[width=0.47\linewidth]{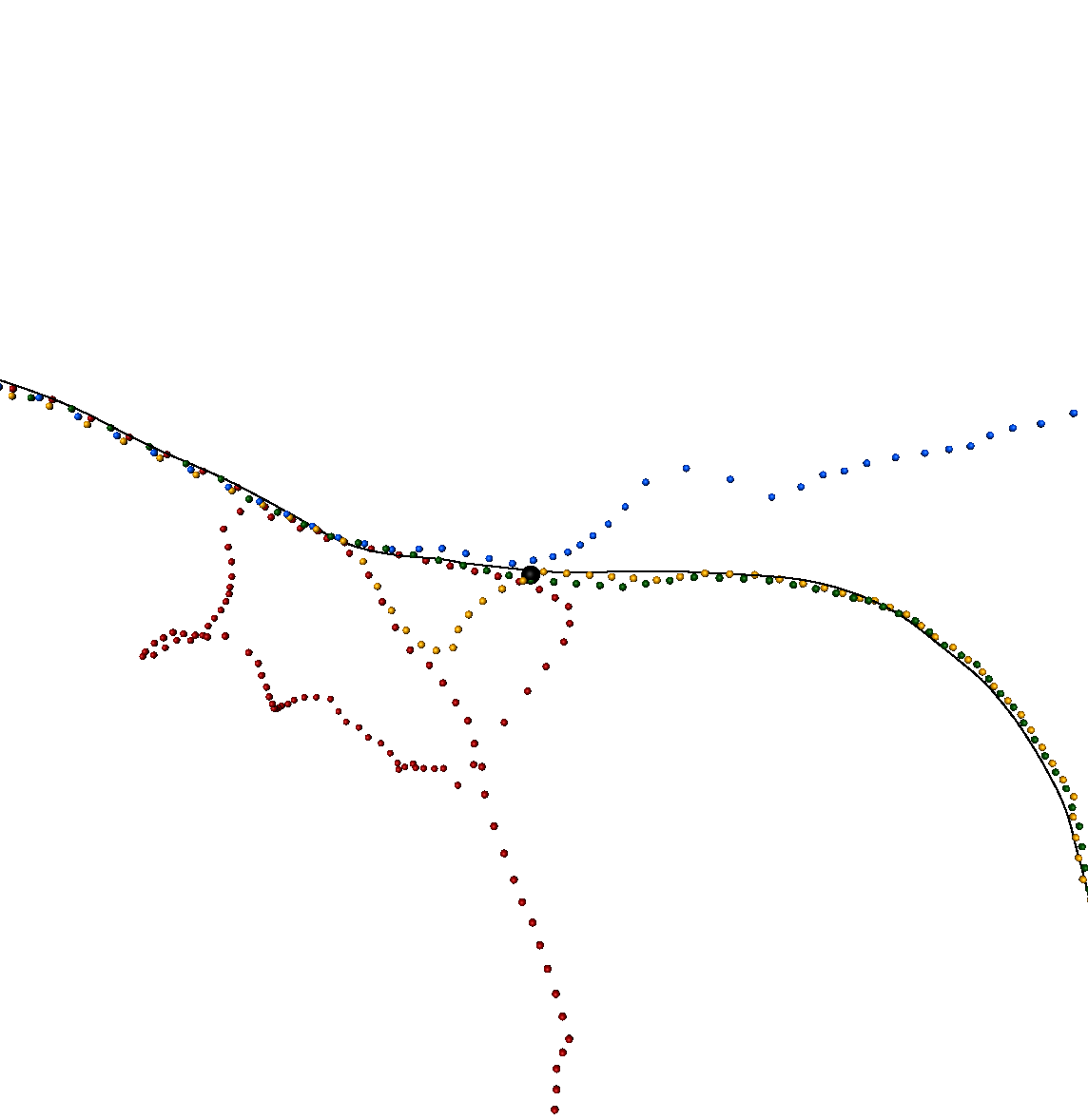}
\label{subfig:notranslation}
} \hfill
\subfloat[Trained with off-centerline points]{
\includegraphics[width=0.47\linewidth]{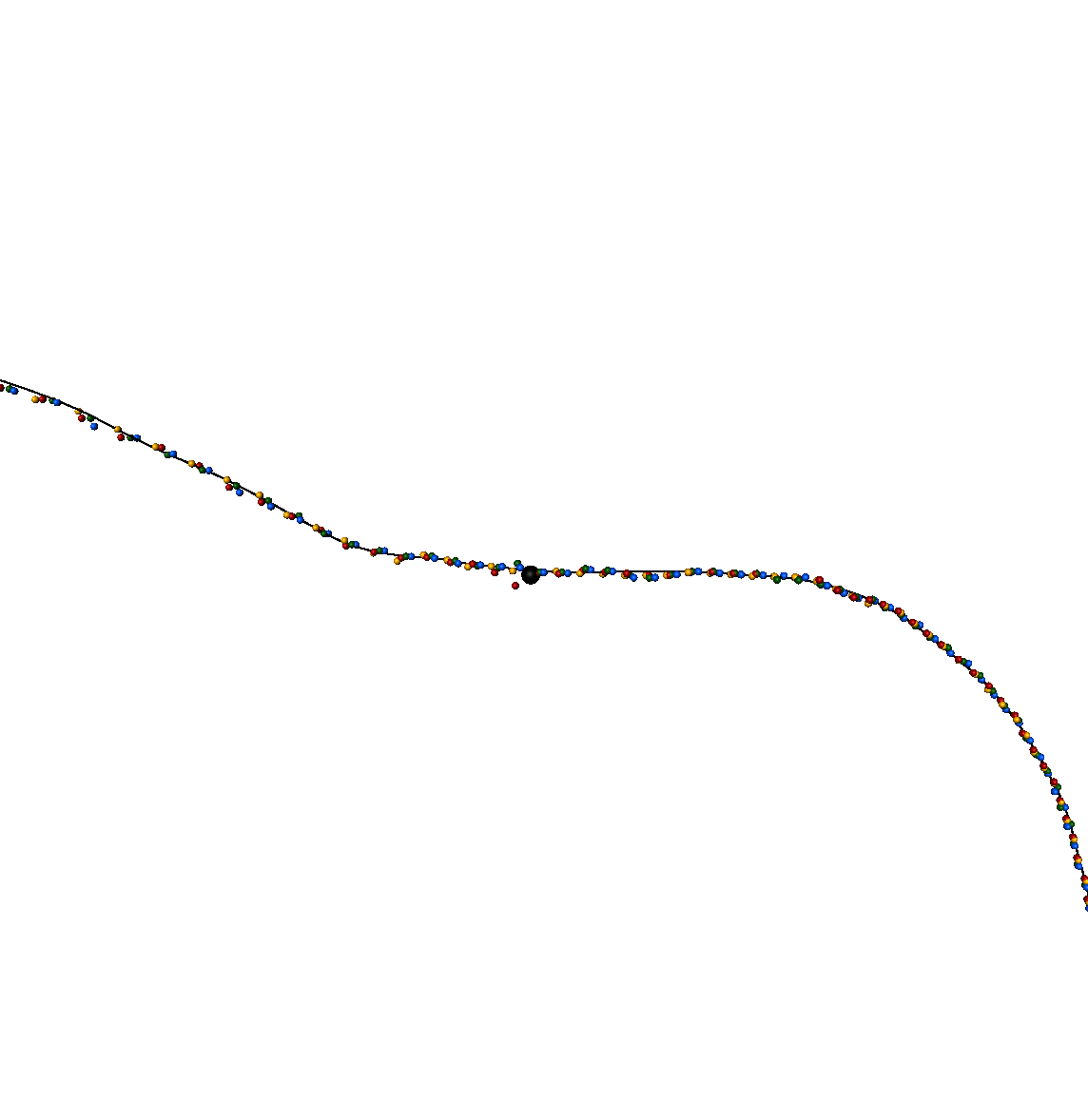}
\label{subfig:withtranslation}
}
\caption{Four extracted centerlines (in blue, red, green, and yellow) starting from random locations around a fixed seed point (black sphere), for \protect\subref{subfig:notranslation} a CNN trained without data augmentation using off-centerline samples, and \protect\subref{subfig:withtranslation} a CNN trained with off-centerline data augmentation. The reference centerline is shown in black. Data augmentation greatly increases the robustness of the method.}
\label{fig:seednoaugment}
\end{figure}

\subsection{Effect of data augmentation}
\label{sec:augment}
The models were trained using rotational and translational data augmentation (Sec. \ref{sec:training}). To evaluate the effect of this data augmentation, we trained additional models without one or both of these augmentation strategies. Fig. \ref{fig:quantaugment} shows the accuracy and overlap of centerlines extracted in the CAT08 training set using different combinations of augmentations during training. Omitting both augmentations affects both overlap and accuracy. Rotation augmentation mostly improves overlap. Translation augmentation improves both overlap and accuracy. Using both types of augmentation substantially both metrics.
In addition, we performed a qualitative analysis of models with and without translation augmentation. For this, we selected a scan and initialized tracking from a seed point. This point was randomly translated multiple times to simulate off-centerline seed points. Fig. \ref{fig:seednoaugment} shows that the CNN trained without translation augmentation is extremely sensitive to seed placement, and unable to recover from deviations from the centerline. In contrast, the CNN trained with translation augmentation immediately recovers the centerline from erroneous seed placement. 
This leads to increased robustness and may allow a higher tolerance in (human) seed point placement.

\section{Discussion}
\label{sec:discussion}
We have presented a deep learning-based method for coronary artery centerline extraction in CCTA. The method does not require hand-crafted filters or features to determine the orientation and size of the coronary artery, but instead uses a CNN to directly extract this information from CCTA images. The CNN can be trained using sparsely annotated vessel trees, which might also make it applicable to other data.

Our experiments showed that the CNN was able to provide a single-vessel seed-based iterative tracker with information about the direction of the coronary centerline and the radius of the coronary lumen at a given point. Experiments showed that this allowed fast tracking of arteries, with an accuracy ranking third among 25 publicly reported results for the 24 test CCTA images of the CAT08 evaluation framework. While the methods ranking first and second required both the starting point and the end point of the coronary artery, our method only requires one seed point, which could be set manually or automatically anywhere in the coronary artery. The tracker automatically identifies the proximal and distal end-point of the centerline based on the entropy in the probability distribution provided by the CNN. The accuracy of the extracted centerlines was high, with an average distance of 0.21 mm to manually annotated reference centerlines. In addition, our experiments showed that the method was able to generalize to images acquired on different scanners, even though the appearance of coronary arteries in CCTA may differ substantially between scanners \citep{Kris13}: A CNN that was trained using the eight CAT08 training images acquired on a Siemens scanner allowed centerline extraction in CCTA scans acquired on GE, Philips and Toshiba scanners.

We found that in the CAT08 data the CNN was able to accurately estimate the radius of vessels within one voxel width, similarly to results reported by \citep{Scha11}. This could serve as an initialization for accurate coronary lumen segmentation, which could be used for stenosis identification and coronary artery volume quantification. An analysis of radius estimation in the UMCU data showed comparable limits of agreement, albeit with a systematic overestimation of the model with respect to manual annotations by an expert observer. Upon closer inspection, this turned out to be most likely caused by differences in radius annotation protocol between the CAT08 dataset and UMCU dataset.

The method was trained and evaluated with images containing different degrees of calcification, intravascular stents, luminal narrowing and noise. Whereas previous methods required specific preprocessing steps for e.g. calcium removal \citep{Ceti15}, we found preprocessing not necessary. Hence, the CNN processed the CCTA data directly, did not require substantial downsampling of images \citep{Yang12} and resulted in a centerline that did not need further refinement, as opposed to \citep{Yang12,Zhen13}. In future work, we will investigate pruning methods to remove venous segments from extracted centerlines \citep{Guls16}, and recurrent neural networks to retain a state for the expected intensity values along the current centerline \citep{Poul17}. 

The presented method is supervised and requires representative training data. Hence, manually annotated reference centerlines are required for a number of training scans. 
Our results show that the method can be trained without exhaustive annotation of all coronary artery centerlines in CCTA training images.
The eight training images used contained annotations for only four coronary artery centerlines, instead of the full coronary tree. We trained the CNN using only samples for which unambiguous labels could be defined, by sparsely sampling training patches around the annotated centerlines. The requirement of only limited and sparsely annotated training data might make the method applicable to other problems, such as extraction of vessels in the liver or the brain. To transfer the method to other applications, a small number of hyperparameters have to be set, which define a trade-off in centerline overlap, centerline accuracy, and computational efficiency. First, patch width $w$ (in voxels) and voxel width $v$ (in mm) determining the physical receptive field of the CNN must be adapted. The architecture listed in Table \ref{tab:network} can easily be extended to accommodate smaller or larger values for $w$, as shown in our experiments in Sec. \ref{sec:resultsumcu}. Second, the number of directions $|D|$ on the sphere should be set. A large number of directions allows prediction with higher granularity of the possible orientations. We found that for coronary centerline extraction, increasing $|D|$ led to higher accuracy of the extracted centerlines, i.e. closer correspondence to the reference centerlines. On the other hand, using too many classes made the network more sensitive to noise. It is expected that setting this value very high may lead to a performance decrease as only few training samples per class are available and the training data is used less effectively.

The ability of the CNN to generalize to new and unseen data was to a large extent due to augmentation of the training set. Rotation augmentation led to substantial invariance of the network to rotations in the input patches. Through off-centerline augmentation the CNN learned orientations towards the centerline as well as to its end-points, allowing the tracker to make corrections in locations where it drifts from the centerline (Fig. \ref{subfig:translate}). Similarly, \citep{Guls16} trained an SVM to predict orientations towards the centerline and its end points, including off-centerline samples during training to obtain more robust predictions. In contrast, centerline regression as proposed \cite{Siro15} would point the tracker towards the centerline, but would not provide the orientation of the vessel.
In spite of rotation and off-centerline augmentation, systematic differences between CCTA images, such as luminal contrast attenuation, remain. In previous work, we showed that blood pool HU values in CCTA can differ substantially among patients \citep{Wolt16}. The attenuation in veins scales accordingly, and therefore veins in one scan may have higher attenuation values than arteries in another scan. This may occasionally lead to confusion between arteries and veins. In future work, additional normalization of domain adaptation \citep{Lafa17} could be used to further maximize the value of the training images. 

This work uses a CNN to predict the most likely direction and radius of an artery at any given point in a CCTA image. This prediction is based on a local image patch which here consisted of $19\times 19\times 19$ voxels. In principle, the space of CNN architectures that can perform this prediction is large. For example, downsampling layers could be included, and hyperparameters such as the number of feature maps per layer could be tuned. We here chose to use an architecture that stacks convolution layers with increasing levels of dilation, following the work of \cite{Yu15}. Dilated convolution kernels can rapidly increase the receptive field of the CNN with few convolution layers, a low number of parameters and no loss of resolution in feature representations. Furthermore, they can be applied in a fully convolutional fashion to not just $19\times 19\times 19$ patches, but to images of any size, as demonstrated in ostia and seed detection. This is an advantage over architectures using downsampling layers such as pooling. Stacked layers with increasing levels of dilated convolutions have previously successfully extended receptive fields in 1D signal processing for audio generation and machine translation \citep{Kalc16,Oord16}, and analysis of 2D images \citep{Yu15,Wolt17}. Dilated convolutions are particularly interesting for 3D data, where the number of parameters typically grows cubically with the size of the input.

In this work, we have shown how CNN predictions can be used to track individual coronary centerlines or full coronary trees. The accuracy and overlap of the proposed method were analyzed using quantitative metrics, while the quality of fully automatic tree extraction was assessed in a qualitative manner. This showed that the automatic method missed on average 1.17 segments per CCTA scan, compared to 1.25 reported by a previous method following a similar analysis \cite{Zhou12}. 
The fully automatic extraction scheme used in this work merges extracted centerlines using a bottom-up approach. In future work, the set of merged centerlines could be further processed to identify bifurcations and individual coronary artery segments. A potential limitation of this bottom-up approach is that the automatic seed identification CNN requires training images in which all coronary arteries have been annotated. In future work we will explore alternative top-down approaches such as multiple hypothesis tracking \citep{Frim10}. It is likely that in this case bifurcation detection could be based on the posterior probability distribution over $|D|$. In practice, even fully automatically extracted coronary artery trees require expert quality control, pruning and adding of coronary branches. 
The interactive extraction method proposed here allows rapid extraction of a coronary artery in 0.4 s based on a single seed point. In contrast to other methods \citep{Frim10,Scha11}, this seed point is only required to be in the coronary artery of interest and does not have to correspond to the most proximal or most distal point, making it easy to initialize centerline extraction.   

\section{Conclusions}
A deep learning-based method for coronary artery centerline extraction has been proposed. The results show that a convolutional neural network can learn to simultaneously determine the direction of coronary artery centerlines and the radius of the coronary lumen with high speed and accuracy. 

\section*{Acknowledgments}
This study was financially supported by the project FSCAD, funded by the Netherlands Organisation for Health Research and Development (ZonMw)  in the framework of the research programme IMDI (Innovative Medical Devices Initiative); project 104003009.

We gratefully acknowledge the support of NVIDIA Corporation with the donation of the Titan Xp GPU used for this research.


\bibliography{literature}
\end{document}